\documentclass[journal]{IEEEtran}
\usepackage{amsmath,amsfonts,amssymb}
\usepackage{algorithmic}
\usepackage{array}
\usepackage{subcaption}
\usepackage{textcomp}
\usepackage{stfloats}
\usepackage{url}
\usepackage{multirow}
\usepackage{verbatim}
\usepackage{graphicx}
\usepackage{booktabs}
\usepackage{tabularx}
\usepackage{hyperref}
\usepackage{float}
\usepackage{tikz}
\usetikzlibrary{positioning, fit, calc, shapes.geometric, arrows.meta, shapes.arrows}
\usepackage{pgfplots}
\pgfplotsset{compat=1.18}

\def\BibTeX{{\rm B\kern-.05em{\sc i\kern-.025em b}\kern-.08em
    T\kern-.1667em\lower.7ex\hbox{E}\kern-.125emX}}
\usepackage{balance}

\title{CLFTv2: Efficient Camera-LiDAR Fusion for Semantic Segmentation via Hierarchical Feature Pyramids}

\author{Toomas~Tahves, Mauro~Bellone, and~Raivo~Sell
\thanks{This research has received funding from the European Union's Horizon 2020 Research and Innovation Programme under grant agreement No.~856602 (Finest Twins), and from the European Union's Horizon Europe Research and Innovation Programme under grant agreement No.~101135988 (PLIADES: AI-Enabled Data Lifecycles Optimization and Data Spaces Integration for Increased Efficiency and Interoperability). \textit{(Corresponding author: Toomas Tahves.)}}
\thanks{Toomas Tahves and Raivo Sell are with the Department of Mechanical and Industrial Engineering, Tallinn University of Technology, Tallinn, Estonia (e-mail: toomas.tahves@taltech.ee).}
\thanks{Mauro Bellone is with the FinEst Centre for Smart Cities, Tallinn University of Technology, Tallinn, Estonia, and also with Universitas Mercatorum, Rome, Italy.}}

\begin{document}

\maketitle

\begin{abstract}
Semantic segmentation for autonomous driving requires reliable detection of vulnerable road users (VRUs) despite heavy class imbalance.
We introduce CLFTv2, a hierarchical camera-LiDAR fusion framework replacing global ViT attention with a Swin-based multi-scale encoder and a lightweight FPN-style residual decoder.
Operating in the 2D perspective domain, CLFTv2 integrates multi-scale geometric cues through shifted-window attention and per-scale residual fusion, avoiding the computational overhead of query-matching decoders.
Across three driving datasets, CLFTv2 consistently improves VRU recall.
On ZOD, CLFTv2‑Large achieves 53.5\% mIoU, improving pedestrian IoU from 35.5\% to 44.9\% over the prior CLFT model.
On Waymo, CLFTv2 reaches 61.7\% mIoU. Additionally, a modality-isolation study suggests ViT's global receptive field yields stronger fusion gains only under dense LiDAR returns.
Compared to a Swin-based Mask2Former adaptation, CLFTv2 requires 1.4$\times$ fewer GFLOPs and delivers 2.2$\times$ higher throughput, while achieving comparable overall accuracy.
These results demonstrate that hierarchical local-attention fusion offers an efficient, scalable alternative to global-attention and query-based decoders for real-time on-vehicle perception in intelligent transportation systems. 
Source code is publicly available.
\end{abstract}


\section{Introduction}
\label{sec:intro}
Semantic segmentation for autonomous driving must reliably identify vulnerable road users (VRUs) whose pixel footprint in perspective images is tiny ($0.3$-$0.4\%$ of pixels) yet critical for safety.
Cameras provide dense color and texture at high spatial resolution but lack metric depth, whereas LiDAR returns 3D geometry but yields point clouds that grow increasingly sparse with distance~\cite{valada2019self, xu2020squeezesegv3}. 
Multi-modal fusion bridges these gaps, but extracting reliable features for distant VRUs requires operating at high spatial resolutions. 
Processing dense, multi-modal sensor streams under the strict latency constraints of autonomous driving requires lightweight architectures to efficiently process multi-modal sensor streams.

Recent vision transformers have advanced multi-modal fusion.
Our prior work, CLFT~\cite{clft1, clft2}, demonstrated that a Vision Transformer (ViT) encoder provides excellent cross-modal context.
However, standard ViTs incur quadratic complexity and maintain a single fixed spatial resolution, making it expensive to generate the multi-scale feature pyramids required for small-object detection~\cite{swin, wang2021pyramid}.

To overcome these limitations, we introduce CLFTv2, a hierarchical camera–LiDAR fusion framework built on the Swin Transformer~\cite{swin}.
Shifted‑window self‑attention restricts computation to local windows, reducing complexity to linear in image resolution, while patch merging produces a four‑stage feature pyramid.
In our framework, two weight-shared Swin encoders process projected 2D camera and LiDAR inputs in parallel.
At each scale, modality‑specific residual convolution units refine the features, which are then fused via element‑wise summation and accumulated through a coarse‑to‑fine FPN‑style decoder.
CLFTv2 operates entirely in the 2D perspective‑projection domain: LiDAR points are projected onto the image plane and encoded as a three‑channel coordinate map, enabling direct use of ImageNet‑pretrained backbones without specialized 3D processing.

The main contributions of this paper are:
\begin{itemize}
\item We propose CLFTv2, a hierarchical Swin-based fusion architecture with a lightweight per-scale residual decoder. By extracting multi-scale geometric cues natively, it avoids the heavy computational overhead of universal query-based decoders when applied to closed-set semantic segmentation.

\item Through a multi-dataset benchmark (ZOD, Waymo, ISEAuto), we demonstrate CLFTv2 achieves competitive accuracy against state-of-the-art query decoders under sparse or pseudo-labeled supervision, while demanding substantially lower GFLOPs and delivering 2.2$\times$ higher throughput.

\item A modality-isolation study reveals a practical trade-off: while Swin's hierarchical structure improves small-object resolution, its local-window attention struggles slightly more than ViT's global receptive field to fully exploit dense, localized LiDAR returns. This suggests hybrid global-local architectures as a direction for future work.

\item We demonstrate that our learned multi-modal representations generalize across differing sensor characteristics and geographical datasets, with cross-dataset initialization accelerating convergence and improving early‑epoch accuracy.
\end{itemize}

\section{Related Work}
\label{sec:related}

\subsection{Camera-LiDAR Fusion for Semantic Segmentation}
\label{subsec:related_fusion}

Multi-modal fusion of camera and LiDAR data has been studied across two broadly distinct paradigms: 3D-native representations and 2D perspective-projection representations.

Early deep fusion work such as PointFusion~\cite{xu2018pointfusion} projected image features onto each raw LiDAR point to augment the 3D point representation for bounding-box estimation.
PointPainting~\cite{vora2020pointpainting} extended this by back-projecting 2D semantic scores from a pretrained image segmentation network onto the point cloud before 3D detection.
More recently, BEVFusion~\cite{bevfusion} unifies multi-sensor data in a bird's-eye-view grid before fusing camera and LiDAR features; UniAD~\cite{planningoriented} builds on BEV for end-to-end multi-task driving.
These 3D-native methods preserve metric depth but require specialised voxelization and sparse-convolution pipelines that preclude direct use of ImageNet-pretrained image backbones.

An alternative paradigm encodes both modalities as spatially aligned 2D image tensors by projecting LiDAR points onto the camera image plane.
This allows the CNN and Transformer image backbones to be applied without modification, at the cost of discarding out-of-plane 3D geometry.
Valada~et~al.~\cite{valada2019self} demonstrated self-supervised adaptation of CNN fusion models under this paradigm; SqueezeSegV3~\cite{xu2020squeezesegv3} applied spatially-adaptive convolutions to projected LiDAR range images for efficient point-cloud segmentation.
TransFusion~\cite{bai2022transfusion} introduced transformer cross-attention between image and LiDAR features in the perspective domain for 3D detection.
Our prior work, CLFT~\cite{clft1, clft2}, applied a ViT encoder to this 2D perspective setting.
CLFTv2 operates in the same paradigm, enabling direct comparison, while replacing the ViT backbone with a hierarchical Swin encoder to address the aforementioned single-scale limitations.
Direct numeric comparison with BEV-based 3D methods is outside the scope of this paper, as the two paradigms discard different geometric information and serve different downstream application constraints.

\subsection{Hierarchical Vision Transformers for Dense Prediction}
\label{subsec:related_swin}

ViT's quadratic attention complexity and single-scale output limit its use for dense prediction; the Swin Transformer~\cite{swin} addressed both.
Shifted-window attention restricts computation to local $M{\times}M$ windows, reducing complexity to linear in image resolution, while Patch Merging layers create a four-stage feature pyramid analogous to the feature hierarchy of ResNets.
SwinV2~\cite{swinv2} subsequently extended this architecture with cosine-similarity attention and log-spaced continuous relative position bias, improving stability at larger model sizes and higher resolutions.
Swin-based backbones have been applied to medical image segmentation~\cite{swin_seg} and monocular depth estimation~\cite{swin_depth}; more directly, Swin serves as the default backbone for both MaskFormer~\cite{cheng2021maskformer} and Mask2Former~\cite{cheng2022masked}.

Feature Pyramid Networks (FPN)~\cite{fpn} established the standard pattern for combining multi-scale backbone features in dense prediction: lateral $1{\times}1$ projections normalize channel dimensions across stages, and top-down upsampling merges coarse semantic and fine spatial information.
CLFTv2's decoder follows this pattern, adding per-scale residual fusion of camera and LiDAR streams before the coarse-to-fine accumulation.
The Dense Prediction Transformer (DPT)~\cite{ranftl2021vision} adapted ViT for dense prediction; CLFT~\cite{clft1, clft2} built on DPT for the camera-LiDAR setting, which CLFTv2 extends by replacing the ViT-DPT stack with the Swin-FPN hierarchy for lower complexity and native multi-scale feature access.

\subsection{Query-Based Universal Segmentation}
\label{subsec:related_maskformer}

MaskFormer~\cite{cheng2021maskformer} reformulated semantic segmentation as a set prediction problem: a fixed bank of learned object queries attend to image features via a standard transformer decoder, each producing a class logit and a binary mask; Hungarian matching assigns queries to ground-truth segments during training.
This cast semantic segmentation as a form of instance-agnostic mask classification, removing the need for per-pixel class assignment heads.
Mask2Former~\cite{cheng2022masked} strengthened this design with masked cross-attention, queries attend only within their predicted foreground region, and a multi-scale deformable attention pixel decoder~\cite{zhu2020deformable}, achieving state-of-the-art results across panoptic, instance, and semantic segmentation benchmarks.

These architectures are designed for open-ended scene decomposition where the number and identity of segments must be inferred per image.
For the closed-set, fixed-class semantic segmentation evaluated in this work, the query-matching mechanism introduces training overhead (Hungarian solver, deep supervision across all decoder layers) that does not directly correspond to a task requirement.
We extend both models to the camera-LiDAR fusion setting using the same per-scale residual fusion method as CLFTv2, and evaluate them as strong baselines.

\begin{figure}[t!]
\centering
\resizebox{1.0\linewidth}{!}{\pgfdeclarelayer{background}
\pgfsetlayers{background,main}

\begin{tikzpicture}[
    scale=0.9,
    font=\sffamily\small,
    >=latex,
    node distance=0.8cm,
    block/.style={rectangle, draw, rounded corners, fill=blue!10, align=center, minimum height=1.0cm, minimum width=2.0cm, inner sep=2pt},
    fusion/.style={rectangle, draw, rounded corners, fill=orange!20, align=center, minimum height=0.9cm, minimum width=2.2cm, inner sep=2pt},
    proj/.style={rectangle, draw, fill=yellow!20, align=center, rounded corners, minimum height=0.7cm, minimum width=1.5cm, inner sep=2pt, font=\sffamily\footnotesize},
    input/.style={rectangle, draw, fill=gray!20, align=center, minimum height=0.8cm, minimum width=1.8cm, inner sep=2pt},
    conn/.style={->, thick, rounded corners},
    skip/.style={->, dashed, thick, rounded corners, color=gray}
]

\node[input] (rgb_in) at (-5.0, 0) {RGB Image\\256x256x3};
\node[input] (lidar_in) at (5.0, 0) {LiDAR Image\\256x256x3};

\node[block, below=0.6cm of rgb_in] (rgb_enc4) {\textbf{Swin Stage 4}\\H/32 (8x8)};
\node[block, below=0.6cm of rgb_enc4] (rgb_enc3) {\textbf{Swin Stage 3}\\H/16 (16x16)};
\node[block, below=0.6cm of rgb_enc3] (rgb_enc2) {\textbf{Swin Stage 2}\\H/8 (32x32)};
\node[block, below=0.6cm of rgb_enc2] (rgb_enc1) {\textbf{Swin Stage 1}\\H/4 (64x64)};

\node[block, below=0.6cm of lidar_in] (lidar_enc4) {\textbf{Swin Stage 4}\\H/32 (8x8)};
\node[block, below=0.6cm of lidar_enc4] (lidar_enc3) {\textbf{Swin Stage 3}\\H/16 (16x16)};
\node[block, below=0.6cm of lidar_enc3] (lidar_enc2) {\textbf{Swin Stage 2}\\H/8 (32x32)};
\node[block, below=0.6cm of lidar_enc2] (lidar_enc1) {\textbf{Swin Stage 1}\\H/4 (64x64)};

\draw[conn] (rgb_in) -- (rgb_enc4);
\draw[conn] (rgb_enc4) -- (rgb_enc3);
\draw[conn] (rgb_enc3) -- (rgb_enc2);
\draw[conn] (rgb_enc2) -- (rgb_enc1);

\draw[conn] (lidar_in) -- (lidar_enc4);
\draw[conn] (lidar_enc4) -- (lidar_enc3);
\draw[conn] (lidar_enc3) -- (lidar_enc2);
\draw[conn] (lidar_enc2) -- (lidar_enc1);


\coordinate (f4_pos) at (0, 0 |- rgb_enc4);
\coordinate (f3_pos) at (0, 0 |- rgb_enc3);
\coordinate (f2_pos) at (0, 0 |- rgb_enc2);
\coordinate (f1_pos) at (0, 0 |- rgb_enc1);

\node[fusion] (fusion4) at (f4_pos) {Fusion Block 4\\(H/32)};
\node[fusion] (fusion3) at (f3_pos) {Fusion Block 3\\(H/16)};
\node[fusion] (fusion2) at (f2_pos) {Fusion Block 2\\(H/8)};
\node[fusion] (fusion1) at (f1_pos) {Fusion Block 1\\(H/4)};

\node[proj, left=0.2cm of fusion1, anchor=east] (proj_rgb1) {1$\times$1 Conv};
\node[proj, left=0.2cm of fusion2, anchor=east] (proj_rgb2) {1$\times$1 Conv};
\node[proj, left=0.2cm of fusion3, anchor=east] (proj_rgb3) {1$\times$1 Conv};
\node[proj, left=0.2cm of fusion4, anchor=east] (proj_rgb4) {1$\times$1 Conv};

\node[proj, right=0.2cm of fusion1, anchor=west] (proj_lidar1) {1$\times$1 Conv};
\node[proj, right=0.2cm of fusion2, anchor=west] (proj_lidar2) {1$\times$1 Conv};
\node[proj, right=0.2cm of fusion3, anchor=west] (proj_lidar3) {1$\times$1 Conv};
\node[proj, right=0.2cm of fusion4, anchor=west] (proj_lidar4) {1$\times$1 Conv};

\draw[conn] (rgb_enc1) -- (proj_rgb1);
\draw[conn] (proj_rgb1) -- (fusion1);

\draw[conn] (lidar_enc1) -- (proj_lidar1);
\draw[conn] (proj_lidar1) -- (fusion1);

\draw[conn] (rgb_enc2) -- (proj_rgb2);
\draw[conn] (proj_rgb2) -- (fusion2);

\draw[conn] (lidar_enc2) -- (proj_lidar2);
\draw[conn] (proj_lidar2) -- (fusion2);

\draw[conn] (rgb_enc3) -- (proj_rgb3);
\draw[conn] (proj_rgb3) -- (fusion3);

\draw[conn] (lidar_enc3) -- (proj_lidar3);
\draw[conn] (proj_lidar3) -- (fusion3);

\draw[conn] (rgb_enc4) -- (proj_rgb4);
\draw[conn] (proj_rgb4) -- (fusion4);

\draw[conn] (lidar_enc4) -- (proj_lidar4);
\draw[conn] (proj_lidar4) -- (fusion4);

\draw[conn] (fusion4) -- node[midway, left, font=\scriptsize] {Up$_{2\times}$} (fusion3);
\draw[conn] (fusion3) -- node[midway, left, font=\scriptsize] {Up$_{2\times}$} (fusion2);
\draw[conn] (fusion2) -- node[midway, left, font=\scriptsize] {Up$_{2\times}$} (fusion1);

\node[block, fill=green!10, below=0.6cm of fusion1] (head_seg) {Segmentation Head\\(Conv$_{3\times3}$-BN-ReLU-Conv$_{1\times1}$)};
\node[input, below=0.6cm of head_seg, fill=white] (output) {Segmentation Mask\\$H\times W$};

\draw[conn] (fusion1.south) -- node[midway, right, font=\scriptsize] {Up$_{4\times}$} (head_seg.north);
\draw[conn] (head_seg) -- (output);

\end{tikzpicture}}
\caption{CLFTv2 Architecture.
Parallel Swin Transformer backbones extract four-level feature pyramids independently from the aligned RGB and LiDAR inputs.
Spatial resolution and channel dimensions are first unified using per-scale $1{\times}1$ projections (yellow) that sit between each backbone stage and the fusion decoder.
The coarse-to-fine residual fusion decoder (orange) propagates semantic context bottom-up: it begins at Stage~4 ($H/32$) and refines the representation via $2{\times}$ bilinear upsampling and stage-wise feature summation through Stages~3, 2, and~1.
The final fused representation $A_1$ ($H/4$) is processed by the segmentation head and upsampled $4{\times}$ to yield the full-resolution categorical prediction.}
\label{fig:architecture}
\end{figure}
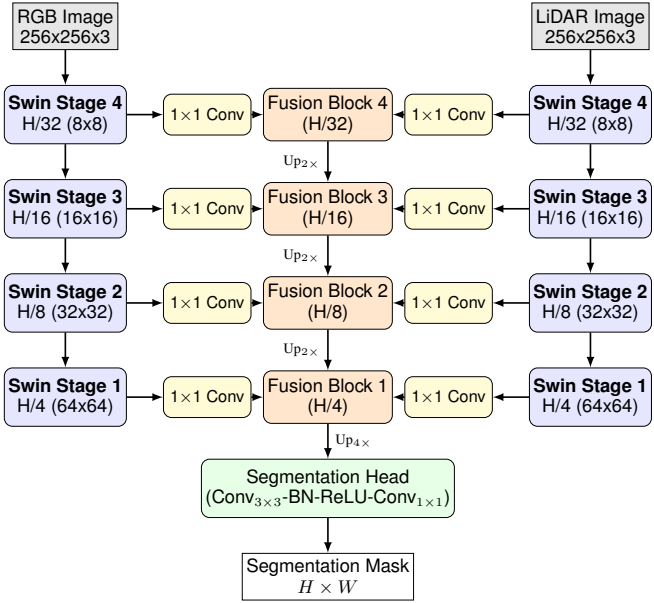

\section{Methodology}
\label{sec:method}

As illustrated in Fig.~\ref{fig:architecture}, CLFTv2 comprises three primary components: a Siamese Swin Transformer encoder that independently extracts hierarchical feature pyramids from aligned RGB and LiDAR inputs; a coarse-to-fine residual fusion decoder that integrates these streams across four spatial scales; and a lightweight segmentation head that upsamples the final fused representation to the original input resolution.
Detailed schematics of the decoder's per-scale fusion mechanism and its core residual convolution units are provided in Fig.~\ref{fig:fusion_module} and Fig.~\ref{fig:resconv_unit}, respectively.

\subsection{Hierarchical Backbone: Swin Transformer}
\label{subsec:swin_backbone}

The input tensor is first partitioned into non-overlapping $4{\times}4$ patches, then processed through four hierarchical transformer stages.
Within each stage, self-attention is computed locally over $M{\times}M$ non-overlapping windows.
For our main benchmarks, the Tiny variant uses a window size of $M{=}16$ at $256{\times}256$ input resolution, while the Base and Large variants natively support scaling $M$ from $12$ to $24$ at $384{\times}384$ resolution (Section~\ref{subsubsec:ablation_window} explicitly investigates varying this window size).
A Shifted-Window (SW-MSA) mechanism in alternating transformer blocks cyclically shifts the window partition by $(\lfloor M/2 \rfloor, \lfloor M/2 \rfloor)$ to allow cross-window information flow.
For a feature map of spatial size $h{\times}w$ and channel dimension $C$, the W-MSA complexity is:
\begin{equation}
\Omega(\text{W-MSA}) = 4hwC^2 + 2M^2hwC,
\end{equation}
which is $\mathcal{O}(hw)$, compared to $\mathcal{O}((hw)^2)$ for global self-attention.
Between stages, Patch Merging halves the spatial resolution and doubles the channel dimension, producing a four-level pyramid $\{F_i\}_{i=1}^{4}$ at strides $S_i \in \{4, 8, 16, 32\}$.
The embedding dimensions $C_i$ scale with model size: $\{96, 192, 384, 768\}$ for Tiny, $\{128, 256, 512, 1024\}$ for Base, and $\{192, 384, 768, 1536\}$ for Large.
In all experiments we use SwinV2 initialized with ImageNet-pretrained weights (ImageNet-1k for Tiny models, ImageNet-22k for Base/Large models).
These pretrained models incorporate cosine-similarity attention and log-spaced continuous relative position bias for improved large-resolution fine-tuning~\cite{swinv2}.

\subsection{Siamese Feature Encoding}
\label{subsec:siamese}

Both the camera image $I_{RGB} \in \mathbb{R}^{H \times W \times 3}$ and the projected LiDAR tensor $I_{LiDAR} \in \mathbb{R}^{H \times W \times 3}$ are processed by two Swin Transformer backbones sharing identical parameters $\theta_{enc}$:
\begin{align}
\{F^{RGB}_i\}_{i=1}^4 &= \mathcal{E}(I_{RGB};\, \theta_{enc}) \\
\{F^{LiDAR}_i\}_{i=1}^4 &= \mathcal{E}(I_{LiDAR};\, \theta_{enc})
\end{align}
where $\mathcal{E}$ denotes the hierarchical encoder mapping a 3-channel input to a four-stage feature pyramid $\{\mathbb{R}^{H_i \times W_i \times C_i}\}_{i=1}^4$, with spatial dimensions $H_i = H/S_i$ and $W_i = W/S_i$ governed by descending strides $S_i \in \{4, 8, 16, 32\}$. 

Rather than visual texture, the LiDAR input encodes structural geometry as a dense coordinate map where each pixel represents a projected 3D point $(X,Y,Z)$.
Formulating this sparse spatial data as a dense 3-channel tensor ensures structural compatibility with the standard vision backbone.
A Siamese architecture with tied weights halves the parameter count and enables the use of ImageNet-pretrained weights across both appearance and geometry streams.

\subsection{Hierarchical Fusion Decoder}
\label{subsec:decoder}

\subsubsection{Channel Projection}
\label{subsubsec:projection}

Each backbone level $i$ produces features $F_{m,i} \in \mathbb{R}^{H_i \times W_i \times C_i}$ with varying embedding dimension $C_i$.
A learned $1{\times}1$ convolution projects each level to a unified dimension $D{=}256$ while preserving the native spatial resolution:
\begin{equation}
F'_{m,i} = \mathrm{Conv}_{1\times1}(F_{m,i}) \in \mathbb{R}^{H_i \times W_i \times D}.
\label{eq:channel_proj}
\end{equation}
This normalises the feature scale across levels and enables consistent fusion operations without any spatial resampling at this stage.

\subsubsection{Coarse-to-Fine Residual Fusion}
\label{subsubsec:fusion_theory}

\begin{figure}[t]
\centering
\resizebox{0.99\linewidth}{!}{\pgfdeclarelayer{background}
\pgfsetlayers{background,main}

\begin{tikzpicture}[
    scale=1,
    font=\sffamily\footnotesize,
    >=latex,
    node distance=0.6cm and 0.8cm,
    block/.style={rectangle, draw, rounded corners, fill=blue!10, align=center, minimum height=0.8cm, minimum width=1.5cm, inner sep=2pt},
    op/.style={circle, draw, fill=yellow!20, minimum size=0.5cm, inner sep=0pt},
    input/.style={rectangle, draw, fill=gray!15, align=center, minimum height=0.6cm, minimum width=1.4cm},
    conn/.style={->, thick, rounded corners},
]

\node[input] (input) at (0, 0) {Input\\$x$};

\node[op, right=0.8cm of input] (relu1) {ReLU};

\node[block, right=0.8cm of relu1] (conv1) {Conv2d\\3x3};

\node[op, right=0.8cm of conv1] (relu2) {ReLU};

\node[block, right=0.8cm of relu2] (conv2) {Conv2d\\3x3};

\node[op, right=0.8cm of conv2] (add) {$+$};

\node[input, fill=orange!20, right=0.8cm of add] (output) {Output};


\draw[conn] (input) -- (relu1);
\draw[conn] (relu1) -- (conv1);
\draw[conn] (conv1) -- (relu2);
\draw[conn] (relu2) -- (conv2);
\draw[conn] (conv2) -- (add);
\draw[conn] (add) -- (output);

\draw[conn] (input.south) -- ++(0,-0.5) -| (add.south);

\end{tikzpicture}}
\caption{Architecture of the Residual Convolution (ResConv) unit.
This module consists of two consecutive $3{\times}3$ convolutions, each followed by a ReLU activation, with an identity skip connection that mirrors standard ResNet topologies~\cite{he2016deep}.}
\label{fig:resconv_unit}
\end{figure}
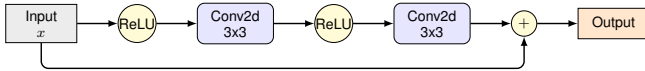

The decoder processes levels in order from the coarsest ($i{=}4$, stride 32) to the finest ($i{=}1$, stride 4), accumulating a fused representation $A_i$ at each scale (see Fig.~\ref{fig:fusion_module}).
Let $\mathcal{R}_{\text{RGB}}$ and $\mathcal{R}_{\text{LiDAR}}$ denote modality-specific Residual Convolution (ResConv) units that independently refine features, and let $\mathcal{R}_{\text{out}}$ denote a shared output ResConv unit applied after fusion (both detailed in Fig.~\ref{fig:resconv_unit}).

At the coarsest level, the accumulation is initialised using only the refined features:
\begin{equation}
A_4 = \mathcal{R}_{\text{out}}\!\left(\mathcal{R}_{\text{RGB}}(F'_{RGB,4}) + \mathcal{R}_{\text{LiDAR}}(F'_{LiDAR,4})\right).
\label{eq:fusion_init}
\end{equation}

At each subsequent finer level $i \in \{3, 2, 1\}$, the accumulated coarser representation is bilinearly upsampled by $2{\times}$ to match the current spatial resolution. This provides top-down semantic context, which is added to the refined modality features before passing through the final output convolution:
\begin{align}
A_i &= \mathcal{R}_{\text{out}}\!\Big(\mathcal{R}_{\text{RGB}}(F'_{RGB,i}) + \mathcal{R}_{\text{LiDAR}}(F'_{LiDAR,i}) \notag \\
    &\qquad \quad + \mathrm{Up}_{2\times}(A_{i+1})\Big).
\label{eq:fusion_recurrence}
\end{align}

The output of the decoder is $A_1 \in \mathbb{R}^{(H/4) \times (W/4) \times D}$.
This coarse-to-fine accumulation combines global semantic context (propagated from $A_4$) with bottom-up fine-grained detail at each level, analogous to the top-down pathway of FPN~\cite{fpn}, while the modality-specific ResConvs align the visual and geometric features prior to linear combination.

\begin{figure}[t]
\centering
\resizebox{1.0\linewidth}{!}{\pgfdeclarelayer{background}
\pgfsetlayers{background,main}

\begin{tikzpicture}[
    scale=0.85,
    font=\sffamily\footnotesize,
    >=latex,
    node distance=0.6cm and 0.8cm,
    block/.style={rectangle, draw, rounded corners, fill=blue!10, align=center, minimum height=0.8cm, minimum width=1.5cm, inner sep=2pt},
    op/.style={circle, draw, fill=yellow!20, minimum size=0.5cm, inner sep=0pt},
    input/.style={rectangle, draw, fill=gray!15, align=center, minimum height=0.6cm, minimum width=1.4cm},
    conn/.style={->, thick, rounded corners},
]

\node[input] (rgb_in) at (0, 1) {Feature\\RGB};
\node[input] (lidar_in) at (0, -1) {Feature\\LiDAR};
\node[input] (prev_in) at (4, -2.5) {$\mathrm{Up}_{2\times}(A_{i+1})$\\(zero if $i{=}4$)};

\node[block, right=0.6cm of rgb_in] (res_rgb) {ResConv\\Unit};
\node[block, right=0.6cm of lidar_in] (res_lidar) {ResConv\\Unit};

\node[op] (sum_mod) at ($(res_rgb)!0.5!(res_lidar)$) [xshift=1.5cm] {$+$};

\node[op, right=0.8cm of sum_mod] (final_sum) {$+$};

\node[block, right=0.8cm of final_sum] (res2) {ResConv\\Unit};

\node[input, fill=orange!20, right=0.6cm of res2] (output) {Output\\$A_i$};


\draw[conn] (rgb_in) -- (res_rgb);
\draw[conn] (lidar_in) -- (res_lidar);

\draw[conn] (res_rgb) -- (sum_mod);
\draw[conn] (res_lidar) -- (sum_mod);

\draw[conn] (sum_mod) -- (final_sum);

\draw[conn] (prev_in.base east) -| (final_sum.south);

\draw[conn] (final_sum) -- (res2);

\draw[conn] (res2) -- (output);

\end{tikzpicture}}
\caption{Detailed architecture of the per-scale fusion block.
Per-modality Residual Convolution (ResConv) units refine the $1{\times}1$ projected features independently.
These refined features are then summed together with the upsampled context accumulated from the preceding, coarser stage.
A final shared ResConv unit processes the combined representation before it is passed to the next finer stage.}
\label{fig:fusion_module}
\end{figure}
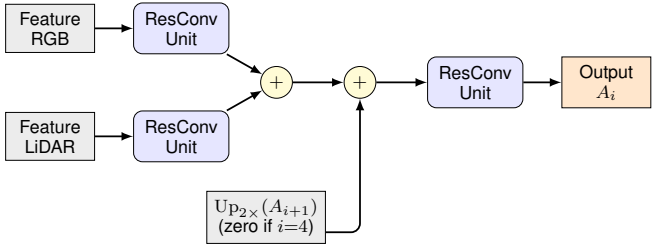

\subsubsection{Ablation Experiments for Design Choices}
\label{subsubsec:ablation_design}

The recurrence term $\mathrm{Up}_{2\times}(A_{i+1})$ in Eq.~\eqref{eq:fusion_recurrence} is the dominant contributor to accuracy.
Removing it (i.e., $A_i = \mathcal{R}_{\text{out}}(\mathcal{R}_{RGB}(F'_{RGB,i}) + \mathcal{R}_{LiDAR}(F'_{LiDAR,i}))$, equivalent to the ResConv Fusion row in Table~\ref{tab:ablation_fusion}) reduces ZOD mIoU from 35.9\% to 23.1\%, an absolute decrease of 12.8\%.
By contrast, replacing element-wise summation with any of five gated or attention-weighted alternatives changes mIoU by at most $\pm$0.7\% across all single-run configurations, confirming that additional fusion parameters are not warranted under the supervision conditions tested.

\subsubsection{Segmentation Head}
\label{subsubsec:segmentation_head}

$A_1$ is at one-quarter of the input resolution.
The segmentation head refines the features with a $3{\times}3$ convolution followed by Batch Normalization (BN) and a ReLU activation, then applies a $1{\times}1$ convolution and upsamples the result to the original resolution:
\begin{equation}
\hat{Y} = \mathrm{Up}_{4\times}\!\left(\mathrm{Conv}_{1\times1}\!\left(\mathrm{ReLU}\!\left(\mathrm{BN}\!\left(\mathrm{Conv}_{3\times3}(A_1)\right)\right)\right)\right)
\end{equation}
producing logits $\hat{Y} \in \mathbb{R}^{H \times W \times C_{cls}}$ where $C_{cls}$ is the number of semantic classes.

\section{Dataset Preparation}
\label{sec:dataset}

\subsection{Datasets}
\label{subsec:datasets}

Three autonomous driving datasets are used in this work, spanning different geographical regions, sensor configurations, annotation methods, and class distributions.
Split sizes and per-class pixel distributions are summarised in Table~\ref{tab:dataset_stats}.

\subsubsection{Waymo Open Dataset}
The Waymo Open Dataset~\cite{waymo} provides large-scale recordings from urban environments across multiple US cities, covering a wide range of weather conditions, occlusion levels, and traffic densities.
We use a 22,000-frame split with front-facing camera and LiDAR captures.
Per-pixel segmentation annotations covering four classes (background, vehicle, human, sign) were prepared within our research group in prior work~\cite{clft1, clft2}; the sensor projection and annotation from that work are unchanged.
Waymo's annotations are dense and manually verified, providing the most reliable ground truth of the three datasets.

\subsubsection{ISEAuto Dataset}
The ISEAuto dataset~\cite{iseauto,clfcn} was collected in Estonian urban and suburban conditions using a vehicle equipped with calibrated camera and LiDAR sensors.
We use a 2,400-frame split containing two foreground semantic classes (human, vehicle).
Unlike ZOD, ISEAuto provides native pixel-level segmentation labels verified by human annotators.
ISEAuto's high foreground class frequency and manually verified annotations make it a complementary evaluation setting to ZOD, directly exposing the effect of annotation quality on model performance.

\subsubsection{Zenseact Open Dataset (ZOD)}
ZOD~\cite{zod} provides multi-modal frames recorded across diverse Swedish urban and highway environments, encompassing challenging scenarios such as snow, rain, and low-light conditions.
We utilize a curated split of 2,300 frames (1,150 for training, 1,150 for validation).
Because ZOD natively supplies only 3D bounding-box annotations, dense four-class segmentation masks were generated via a semi-automated SAM-based~\cite{sam} projection pipeline. 
This process relied on manually selected frames and a suite of algorithmic quality filters (size-filtering, priority-based fusion, sequence-consistency). While this human-in-the-loop curation ensures the masks are not purely unsupervised, they remain pseudo-labels subject to minor local inaccuracies.
A specific consequence of this generation pipeline is a higher mask rejection rate for occluded or heavily truncated objects. ZOD IoU scores should be interpreted as an upper-bound estimate constrained by geometric visibility rather than pure sensor-level detection capability.

\begin{table}[htbp]
\centering
\caption{Dataset Pixel Statistics by Train, Validation and Test Split. Values show percentage of total pixels per class; ISEAuto does not include Sign class annotations (denoted --).}
\begin{tabular}{@{}l l r r r r r@{}}
\toprule
Dataset & Split & Frames & Background & Vehicle & Sign & Human \\
\midrule
\multirow{3}{*}{ZOD}
 & Train      & 1,148  & 97.8\% & 1.5\% & 0.4\% & 0.3\% \\
 & Val. &    573 & 97.7\% & 1.6\% & 0.4\% & 0.2\% \\
 & Test       &    579 & 97.8\% & 1.5\% & 0.4\% & 0.3\% \\
\midrule
\multirow{3}{*}{Waymo}
 & Train      & 13,199 & 95.5\% & 4.1\% & 0.1\% & 0.3\% \\
 & Val. &  4,400 & 95.4\% & 4.1\% & 0.1\% & 0.3\% \\
 & Test       &  4,400 & 95.3\% & 4.2\% & 0.1\% & 0.3\% \\
\midrule
\multirow{3}{*}{ISEAuto}
 & Train      &  1,200 & 96.7\% & 2.9\% & --  & 0.3\% \\
 & Val. &    600 & 96.5\% & 3.1\% & --  & 0.3\% \\
 & Test       &    600 & 96.8\% & 2.9\% & --  & 0.4\% \\
\bottomrule
\end{tabular}
\label{tab:dataset_stats}
\end{table}

\subsection{Data Representation and Preprocessing}
We transform the raw camera and LiDAR data into spatially aligned tensors that can be processed by our hierarchical encoder.

\subsubsection{Input Preprocessing}
Camera images are resized to $H{\times}W$ and normalized with standard ImageNet statistics ($\boldsymbol{\mu}=(0.485,0.456,0.406)$, $\boldsymbol{\sigma}=(0.229,0.224,0.225)$), consistent with all pretrained backbones used.

Raw LiDAR points are projected onto the 2D image plane using the sensor's extrinsic and intrinsic calibration parameters. For each pixel, the spatial coordinates $(x,y,z)$ of the closest projected return are stored in a three-channel format, natively substituting the standard RGB channels (where $Z$ provides depth).
To store these sparse maps efficiently as 8-bit PNGs, we apply dataset-specific quantization rules. 
For ZOD and ISEAuto, the coordinate distributions are clipped at the 95th percentile, scaled to $[-1, 1]$, and mapped into the integer range $[0, 255]$, assigning empty pixels to a neutral background value of 127. 
Conversely, for the Waymo dataset, min-max normalization is applied dynamically per-frame to fill the standard $[0, 255]$ bounds. 
During training, these integer maps are re-normalized into a continuous floating-point format: $I_{LiDAR}=(I_{PNG}/255 - \boldsymbol{\mu}_{LiDAR})/\boldsymbol{\sigma}_{LiDAR}$, where dataset-specific statistics ($\boldsymbol{\mu}_{LiDAR}$, $\boldsymbol{\sigma}_{LiDAR}$) are computed across the active training split (Table~\ref{tab:lidar_stats}).
While these different scaling approaches mean the numerical value of empty pixels fluctuates per-frame, the networks successfully learn to distinguish these background artifacts from true geometric foreground structures regardless of the dataset scheme.

\begin{table}[h]
\centering
\caption{Dataset-specific LiDAR PNG normalization statistics ($\boldsymbol{\mu}_{LiDAR}$, $\boldsymbol{\sigma}_{LiDAR}$) computed over non-zero pixels in the float $[0,1]$ domain after 8-bit decoding.
Channels correspond to R=X (left-right), G=Y (up-down), B=Z (depth).}
\label{tab:lidar_stats}
\setlength{\tabcolsep}{5pt}
\begin{tabular}{lcccccc}
\toprule
 & \multicolumn{3}{c}{$\boldsymbol{\mu}_{LiDAR}$ (R, G, B)} & \multicolumn{3}{c}{$\boldsymbol{\sigma}_{LiDAR}$ (R, G, B)} \\
\cmidrule(lr){2-4}\cmidrule(lr){5-7}
Dataset & R & G & B & R & G & B \\
\midrule
Waymo   & 0.461 & 0.288 & 0.267 & 0.115 & 0.126 & 0.098 \\
ZOD     & 0.253 & 0.493 & 0.496 & 0.234 & 0.031 & 0.167 \\
ISEauto & 0.824 & 0.430 & 0.510 & 0.170 & 0.339 & 0.348 \\
\bottomrule
\end{tabular}
\end{table}

\subsection{Class Weights}
\label{subsec:training}

We train with Weighted Cross-Entropy Loss:
\begin{equation}
\mathcal{L}_{WCE} = -\frac{1}{N} \sum_{n=1}^{N} \sum_{c=1}^{C} w_c \cdot y_{n,c} \cdot \log \hat{y}_{n,c}
\end{equation}
where $N$ is the number of pixels, $y_{n,c} \in \{0,1\}$ is the ground-truth indicator, $\hat{y}_{n,c}$ is the predicted probability, and $w_c$ is the class-specific weight.
Weights are derived per dataset from the per-class pixel frequency using an approximate inverse-square-root schedule, as detailed in Table~\ref{tab:class_weights}.
The low background weight suppresses the dominant gradient contribution of non-object pixels.

\begin{table}[h]
\centering
\caption{Class Weights per Dataset}
\label{tab:class_weights}
\begin{tabular}{l *{4}{c}}
\toprule
Dataset & Background & Vehicle & Human & Sign \\
\midrule
Waymo & 0.5 & 4.0 & 10.0 & 10.0 \\
ZOD & 0.1 & 10.0 & 20.0 & 17.0 \\
ISEAuto & 0.1 & 10.0 & 20.0 & - \\
\bottomrule
\end{tabular}
\end{table}

\section{Experiments and Results}
\label{sec:exp_results}

\subsection{Experimental Setup}
\label{subsec:setup}
\subsubsection{Training Configuration}
All models were implemented in PyTorch and trained on NVIDIA A100 GPUs (80GB) provided by the TalTech High Performance Computing Centre~\cite{herrmann} with a total training time of over 800 hours.
Training configurations (optimizers, learning rate schedules, total epochs, and input resolutions) were tailored to match the optimal setup for each architecture baseline (Table~\ref{tab:hyperparameters}).
A common batch size of $8$ and a base learning rate of $8{\times}10^{-5}$ were preserved across all setups.
General data augmentation across setups included random horizontal flipping ($p{=}0.5$), rotation ($\pm 20^\circ$, $p{=}0.4$), and random resized cropping ($p{=}0.3$).

\begin{table}[h]
\centering
\caption{Training configurations across models and datasets.}
\label{tab:hyperparameters}
\resizebox{\columnwidth}{!}{%
\begin{tabular}{l c c l}
\toprule
\multirow{2}{*}{Architecture} & \multicolumn{2}{c}{Total Epochs} & \multirow{2}{*}{Resolution} \\
\cmidrule(lr){2-3}
 & Waymo & ZOD / ISEAuto & \\
\midrule
CLFTv2 (Base / Large) & 100 & \multicolumn{1}{c}{200} & $384{\times}384$ \\
CLFTv2 (Tiny) & 100 & \multicolumn{1}{c}{200} & $256{\times}256$ \\
CLFTv1~\cite{clft1} & 200 & \multicolumn{1}{c}{300} & $384{\times}384$ \\ 
DeepLabV3+ & 100 & \multicolumn{1}{c}{200} & $256{\times}256$ \\
MaskFormer (Base / Large) & \hphantom{0}50 & \multicolumn{1}{c}{100} & $384{\times}384$ \\
MaskFormer (Tiny) & \hphantom{0}50 & \multicolumn{1}{c}{100} & $256{\times}256$ \\
Mask2Former (Base / Large) & \hphantom{0}50 & \multicolumn{1}{c}{100} & $384{\times}384$ \\
Mask2Former (Tiny) & \hphantom{0}50 & \multicolumn{1}{c}{100} & $256{\times}256$ \\
\bottomrule
\end{tabular}
}
\end{table}

Although all models use identical seeds and dataset pipelines, training schedules differ to match architecture-specific convergence.
Some models (e.g., MaskFormer variants) plateau earlier, so we use fewer epochs to avoid unnecessary compute.
CLFT results reported here come from the unified codebase and supersede prior publications~\cite{clft1, clft2}.

\begin{table*}[h]
\centering
\caption{Deviations of our MaskFormer and Mask2Former re-implementations from the published originals~\cite{cheng2021maskformer,cheng2022masked}.}
\label{tab:baseline_deviations}
\small
\begin{tabularx}{\linewidth}{@{}p{0.5cm}lXXl@{}}
\toprule
\textbf{\#} & \textbf{Aspect} & \textbf{Original} & \textbf{Our adaptation} & \textbf{Model} \\
\midrule
(i) & CUDA kernel
  & Hand-written C++/CUDA extension~\cite{zhu2020deformable}
  & Pure-PyTorch via \texttt{F.grid\_sample}; numerically equivalent for 2D bilinear sampling.
  & M2F only \\
\addlinespace
(ii) & Backbone
  & Swin-T/S/B~\cite{swin} pretrained on ImageNet-1k/22k
  & SwinV2~\cite{swinv2} pretrained on ImageNet-22k; cosine-similarity attention and log-spaced relative position bias.
  & Both \\
\addlinespace
(iii) & Input modality
  & Single RGB stream
  & Camera + LiDAR dual stream; residual conv units with element-wise addition per backbone scale before the pixel decoder.
  & Both \\
\addlinespace
(iv) & Background in matching
  & Class~0 excluded from GT segments in Hungarian matching
  & Class~0 included as explicit GT segment; motivated by ${>}95\%$ background pixel frequency.
  & Both \\
\addlinespace
(v) & Backbone LR
  & Single LR for all parameters
  & Backbone at $0.1{\times}$ base rate; decoder and head at full base rate.
  & MF only \\
\addlinespace
\bottomrule
\end{tabularx}
\end{table*}

\subsubsection{Baseline Implementations: MaskFormer and Mask2Former}
\label{subsubsec:maskformer_impl}
To reduce framework-induced variance, both MaskFormer~\cite{cheng2021maskformer} and Mask2Former~\cite{cheng2022masked} were re-implemented natively in PyTorch, removing external framework dependencies (e.g., Detectron2) in favor of a simpler, pure PyTorch approach.
These implementations follow the original architectures while incorporating the adaptations listed in Table~\ref{tab:baseline_deviations}.
MaskFormer reproduces the FPN pixel decoder and the six-layer cross-attention transformer decoder ($d_\text{model}{=}256$, 8 heads, $Q{=}100$ queries).
We apply identical Hungarian matching cost weights ($\lambda_\text{cls}{=}1$, $\lambda_\text{focal}{=}20$, $\lambda_\text{dice}{=}1$) and training loss scaling ($\lambda_\text{cls}{=}2$, $\lambda_\text{mask}{=}5$, $\lambda_\text{dice}{=}5$) with deep supervision enabled.
The final segmentation map is assembled using the reference semantic inference formula, mapping predicted class probabilities $\mathbf{c}_q$ and mask logits $m_q$:
\begin{equation}
\hat{s}_c = \sum_{q} \text{softmax}(\mathbf{c}_q)_c \cdot \sigma(m_q)
\end{equation}
where the no-object column is dropped without gating.

Mask2Former extends this paradigm via six transformer encoder layers of multi-scale deformable self-attention, and a nine-layer masked cross-attention decoder with auxiliary sequence losses.
To avoid custom CUDA dependencies, the multi-scale deformable attention is implemented natively in PyTorch; this gives mathematical equivalence to the original custom CUDA kernel~\cite{zhu2020deformable} for 2D bilinear sampling.
Both baseline models natively ingest the identical Camera-LiDAR early-fusion pipeline utilized by CLFTv2 (residual convolutions and element-wise addition across all backbone scales prior to the pixel decoder).
Consequently, comparisons in this paper target the adapted implementations within a common codebase, not direct leaderboard replication of the original single-RGB formulations.

\subsubsection{DeepLabV3+ Baseline Implementation}
To establish a CNN-based comparative baseline, DeepLabV3+~\cite{deeplabv3plus} was incorporated.
The implementation reconstructs the standard architecture: a ResNet-101 backbone pretrained on ImageNet-1k, an Atrous Spatial Pyramid Pooling (ASPP) module with four parallel heads at dilation rates $\{6, 12, 18\}$ alongside global-average-pooling, and a standard decoder.
This decoder upsamples the ASPP output to $\tfrac{1}{4}$ resolution, concatenates it with low-level layer-1 features ($1{\times}1$ projected to 48 channels), and applies two $3{\times}3$ convolution blocks followed by a $1{\times}1$ classifier.
For multimodal evaluation, Camera-LiDAR fusion employs the exact same residual-averaging formulation as CLFTv2: two independent modality-specific ResNet-101 streams are processed, and their parallel ASPP representations are summed prior to the decoder.

\subsubsection{Evaluation Metrics}
\label{subsubsec:metrics}
We evaluate segmentation performance using the Intersection-over-Union (IoU) metric.
Because autonomous driving scenes exhibit extreme class imbalance (often $>95\%$ background), standard mIoU is easily skewed.
To prevent the static background from inflating aggregate scores, we adopt a Foreground mIoU ($\text{mIoU}_{fg}$) which focuses exclusively on dynamic agents:
\begin{equation}
\text{mIoU}_{fg} = \frac{1}{|C_{fg}|} \sum_{c \in C_{fg}} \text{IoU}_c
\end{equation}
where $C_{fg}$ consists of the \textit{vehicle}, \textit{human}, and \textit{sign} classes.

We complement this with Frequency-Weighted Foreground IoU (FW~IoU), weighting each class by its pixel frequency to balance common object validation against improvements on sparse classes~\cite{fcn}.

Finally, to reduce single-epoch noise while avoiding overfitting to a single peak checkpoint, we report metrics as the average of the top $10$ validation checkpoints from each session rather than a single peak score.
This top-10 averaging is applied consistently across all models; we report dispersion ($\pm$) where available.

\subsection{Quantitative Analysis}

\begin{table}[t!]
\centering
\caption{Fusion Results on ZOD mIoU (\%).}
\label{tab:main_results_zod}
\resizebox{\columnwidth}{!}{%
\begin{tabular}{@{}lcccccc@{}}
\toprule
Method & mIoU & mRec & mPrec & Veh IoU & Hum IoU & Sig IoU \\
\midrule
CLFTv2-Large      & \textbf{53.54} & 81.17 & 60.08 & \textbf{72.57} & \textbf{44.90} & \textbf{43.15} \\
MaskFormer-Large  & 52.83 & 64.70 & \textbf{72.48} & 72.18 & 44.33 & 41.98 \\
Mask2Former-Large & 52.52 & 64.19 & 72.37 & 72.27 & 43.26 & 42.04 \\
CLFTv2-Base       & 52.45 & 80.87 & 58.78 & 71.77 & 43.37 & 42.20 \\
MaskFormer-Base   & 52.34 & 64.43 & 71.93 & 71.64 & 43.50 & 41.88 \\
Mask2Former-Base  & 52.07 & 64.14 & 71.72 & 71.31 & 43.28 & 41.63 \\
CLFT-Large        & 46.82 & 65.89 & 59.31 & 62.39 & 35.52 & 33.15 \\
CLFT-Hybrid       & 45.79 & 72.68 & 53.65 & 66.90 & 36.03 & 34.45 \\
CLFT-Base         & 44.64 & 67.30 & 54.64 & 66.58 & 33.66 & 33.69 \\
CLFTv2-Tiny       & 43.44 & 78.06 & 48.32 & 65.34 & 34.10 & 30.89 \\
Mask2Former-Tiny  & 43.28 & 54.16 & 65.32 & 65.18 & 33.24 & 31.43 \\
MaskFormer-Tiny   & 42.90 & 53.44 & 65.68 & 64.44 & 32.93 & 31.33 \\
DeepLabV3+        & 36.70 & \textbf{91.17} & 37.83 & 60.20 & 29.10 & 20.80 \\
\bottomrule
\end{tabular}
}
\end{table}

\subsubsection{Overall Segmentation Performance}
We evaluate CLFTv2 against baselines on ZOD, Waymo, and ISEAuto using architecture-appropriate resolutions and a shared batch-size regime.
On ZOD (Table~\ref{tab:main_results_zod}), CLFTv2-Large reaches 53.5\% mIoU (vs. 46.8\% for CLFT-Large), with large gains on Human IoU (35.5\%\,$\rightarrow$\,44.9\%) and Sign IoU (33.2\%\,$\rightarrow$\,43.2\%).
Mask2Former-Large reaches 52.5\% mIoU, within 1\% of CLFTv2-Large, but with roughly double training time.
Ranking is dataset-dependent: query-based models lead on ISEAuto (MaskFormer-Large 75.6\%, Mask2Former-Large 75.4\%, CLFTv2-Large 73.3\%), while ZOD favors CLFTv2-Large.

\begin{table}[t!]
\centering
\caption{Fusion Results on WAYMO mIoU (\%).}
\label{tab:main_results_waymo}
\resizebox{\columnwidth}{!}{%
\begin{tabular}{@{}lcccccc@{}}
\toprule
Method & mIoU & mRec & mPrec & Veh IoU & Hum IoU & Sig IoU \\
\midrule
CLFT-Large        & \textbf{68.26} & 91.59 & \textbf{72.31} & 79.75 & \textbf{64.88} & 54.99 \\
CLFT-Base         & 66.32 & 91.93 & 69.89 & \textbf{79.80} & \textbf{64.88} & 54.29 \\
CLFT-Hybrid       & 65.50 & 92.25 & 68.87 & 78.10 & 63.30 & \textbf{55.11} \\
CLFTv2-Large      & 61.69 & \textbf{95.30} & 63.49 & 67.76 & 62.60 & 54.73 \\
CLFTv2-Base       & 61.02 & 95.20 & 62.80 & 67.54 & 61.79 & 53.73 \\
CLFTv2-Tiny       & 55.61 & 93.92 & 57.45 & 64.56 & 55.06 & 47.21 \\
MaskFormer-Large  & 50.79 & 65.57 & 69.06 & 56.84 & 50.82 & 44.71 \\
MaskFormer-Base   & 50.39 & 64.77 & 69.22 & 56.88 & 50.39 & 43.90 \\
Mask2Former-Large & 49.30 & 64.06 & 68.01 & 54.14 & 49.41 & 44.34 \\
Mask2Former-Base  & 49.10 & 63.66 & 68.07 & 53.99 & 49.37 & 43.93 \\
DeepLabV3+        & 48.15 & 91.01 & 50.08 & 61.88 & 46.19 & 36.40 \\
MaskFormer-Tiny   & 42.71 & 56.70 & 62.92 & 53.17 & 41.92 & 33.05 \\
Mask2Former-Tiny  & 41.85 & 55.99 & 61.73 & 51.54 & 40.79 & 33.21 \\
\bottomrule
\end{tabular}
}
\end{table}
\begin{table}[t!]
\centering
\caption{Fusion Results on ISEAuto mIoU (\%).}
\label{tab:main_results_iseauto}
\resizebox{\columnwidth}{!}{%
\begin{tabular}{@{}lccccc@{}}
\toprule
Method & mIoU & mRec & mPrec & Veh IoU & Hum IoU \\
\midrule
MaskFormer-Large  & \textbf{75.58} & 85.23 & 86.77 & \textbf{81.15} & \textbf{70.01} \\
Mask2Former-Large & 75.37 & 85.28 & 86.42 & 81.06 & 69.67 \\
MaskFormer-Base   & 75.21 & 84.66 & \textbf{86.84} & 81.02 & 69.39 \\
Mask2Former-Base  & 74.61 & 85.19 & 85.47 & 80.77 & 68.46 \\
CLFTv2-Large      & 73.30 & 94.29 & 76.59 & 79.14 & 67.45 \\
CLFTv2-Base       & 73.23 & 94.03 & 76.69 & 79.18 & 67.27 \\
Mask2Former-Tiny  & 70.13 & 80.80 & 83.74 & 77.48 & 62.78 \\
MaskFormer-Tiny   & 69.96 & 81.09 & 83.23 & 76.71 & 63.20 \\
CLFTv2-Tiny       & 69.94 & 94.24 & 72.95 & 76.49 & 63.39 \\
CLFT-Hybrid       & 69.37 & 93.51 & 72.73 & 75.72 & 63.02 \\
CLFT-Large        & 69.85 & 93.01 & 73.55 & 76.17 & 63.52 \\
CLFT-Base         & 65.67 & 93.78 & 68.46 & 72.89 & 58.46 \\
DeepLabV3+        & 56.97 & \textbf{95.53} & 58.35 & 66.60 & 47.33 \\
\bottomrule
\end{tabular}
}
\end{table}

\begin{table}[t]
\centering
\caption{Window-size ablation: fusion validation mIoU (\%).}
\label{tab:ablation_window}
\begin{tabular}{lcccc}
\toprule
Dataset / Model & Window & Val mIoU & FLOPs (G) & Img. Size \\
\midrule
ISEAuto / Base & 24 & 73.23 & 119.0 & 384 \\
ISEAuto / Base & 16 & 72.05 & 52.7 & 256 \\
ISEAuto / Base & 8  & 70.66 & 52.2 & 256 \\
ISEAuto / Tiny & 16 & 69.94 & 30.9 & 256 \\
ISEAuto / Tiny & 8  & 69.12 & 30.8 & 256 \\
ZOD / Base     & 24 & 52.45 & 119.0 & 384 \\
ZOD / Base     & 16 & 45.72 & 52.7 & 256 \\
ZOD / Base     & 8  & 44.69 & 52.2 & 256 \\
ZOD / Tiny     & 16 & 43.44 & 30.9 & 256 \\
ZOD / Tiny     & 8  & 41.86 & 30.8 & 256 \\
Waymo / Base   & 24 & 61.02 & 119.0 & 384 \\
Waymo / Base   & 16 & 57.01 & 52.7 & 256 \\
Waymo / Base   & 8  & 56.50 & 52.2 & 256 \\
Waymo / Tiny   & 16 & 55.61 & 30.9 & 256 \\
Waymo / Tiny   & 8  & 55.18 & 30.8 & 256 \\
\bottomrule
\end{tabular}
\end{table}

\subsubsection{The Impact of Global vs. Local Attention in Cross-Modal Fusion}
An architectural divergence occurs on the Waymo dataset (Table~\ref{tab:main_results_waymo}), where the ViT-based CLFT family completely outperforms CLFTv2-Large (61.7\% mIoU).
Specifically, CLFT-Large (68.3\%) and CLFT-Base (66.3\%) outpace the Swin-based architecture by 6.6\% and 4.6\% respectively.

Because both architectures achieve nearly identical mIoU (within 0.2--0.4\%) when restricted to single-modality inputs (Table~\ref{tab:ablation_modality}), this gap isolates the fusion interaction mechanism.
ViT's unrestricted global self-attention computes long-range cross-modal associations across the entire dense Waymo LiDAR point cloud.
In contrast, Swin's local window bounds these associations.
While reducing the window size from 24 to 8 degrades accuracy further (Table~\ref{tab:ablation_window}), even the maximum expanded window fails to fully bridge the fusion-efficiency gap against pure global attention on Waymo's dense geometric point clouds.
This confirms that while Swin provides efficiency, unbounded global attention is superior for dense spatial point-cloud alignment.

\subsection{Ablation Studies}
\label{subsec:ablation}
We ablate three factors: Swin window size, modality contribution, and fusion design.

\subsubsection{Window Size Ablation}
\label{subsubsec:ablation_window}
Table~\ref{tab:ablation_window} indicates a context-efficiency trade-off.
For CLFTv2-Base, moving from window 24 to 16 reduces mIoU by $1.18\%$ (ISEAuto), $6.73\%$ (ZOD), and $4.01\%$ (Waymo), while reducing FLOPs from 119.0G to 52.7G.
The additional reduction from window 16 to 8 is smaller: $1.39\%$ (ISEAuto), $1.03\%$ (ZOD), and $0.51\%$ (Waymo) for Base, and $0.82\%$ (ISEAuto), $1.58\%$ (ZOD), and $0.43\%$ (Waymo) for Tiny.
This is not a clean window-only ablation: window-24 runs use 384$\times$384 inputs, whereas window-16/8 runs use 256$\times$256.
Therefore, the observed gap reflects a joint window-size and resolution effect.

\begin{figure*}[h]
    \centering
    \begin{subfigure}[b]{0.32\textwidth}
        \centering
        \caption{RGB-Only}
        \includegraphics[width=\textwidth]{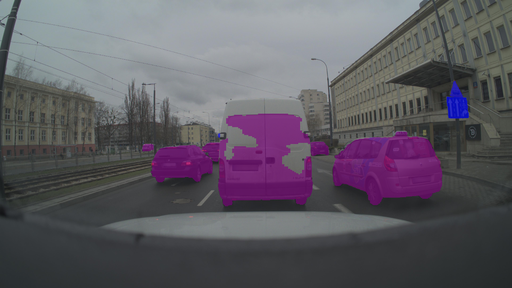}
        \label{fig:rgb_only}
    \end{subfigure}
    \hfill
    \begin{subfigure}[b]{0.32\textwidth}
        \centering
        \caption{LiDAR-Only}
        \includegraphics[width=\textwidth]{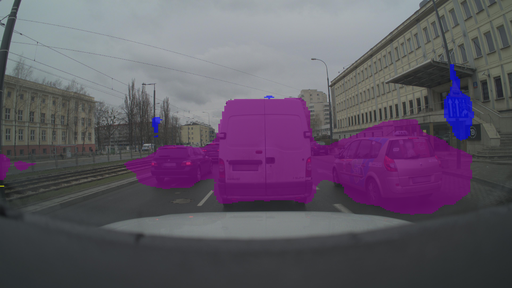}
        \label{fig:lidar_only}
    \end{subfigure}
    \hfill
    \begin{subfigure}[b]{0.32\textwidth}
        \centering
        \caption{Residual Fusion}
        \includegraphics[width=\textwidth]{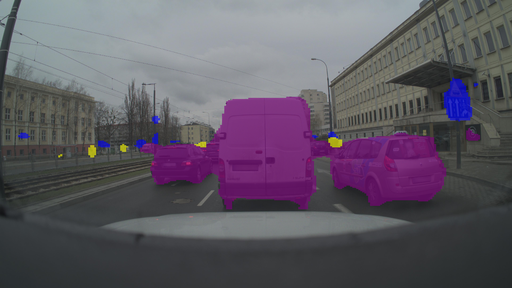}
        \label{fig:fusion_mod}
    \end{subfigure}
    \caption{Modality ablation on ZOD, illustrating failure modes of individual sensors.
    The scene shows a low-contrast white vehicle.
    (a) RGB-only fails to separate the sky and roof of vehicle.
    (b) LiDAR-only recovers shape geometry but hallucinates false classes.
    (c) Residual fusion correctly joins optical boundaries with sparse depth.}
    \label{fig:ablation_modality}
\end{figure*}

\begin{figure*}[h]
    \centering
    \begin{subfigure}[b]{0.32\textwidth}
        \centering
        \caption{Ground Truth}
        \includegraphics[width=\textwidth]{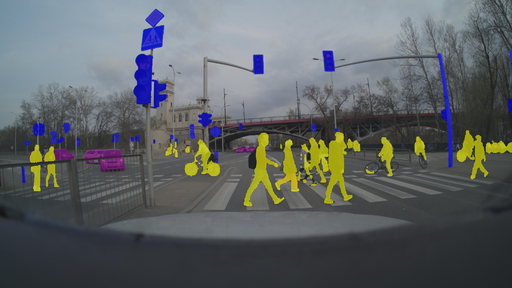}
        \label{fig:gt}
    \end{subfigure}
    \hfill
    \begin{subfigure}[b]{0.32\textwidth}
        \centering
        \caption{Simple Average Fusion}
        \includegraphics[width=\textwidth]{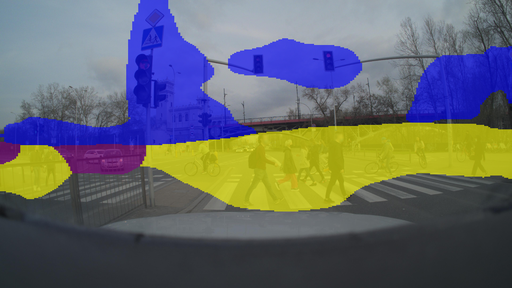}
        \label{fig:simple_avg}
    \end{subfigure}
    \hfill
        \begin{subfigure}[b]{0.32\textwidth}
        \centering
        \caption{Residual Fusion}
        \includegraphics[width=\textwidth]{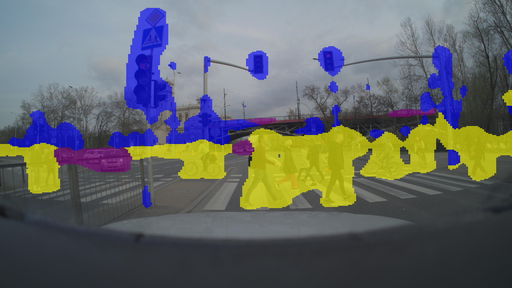}
        \label{fig:residual_fusion}
    \end{subfigure}
    \caption{Visual comparison of fusion strategies using CLFTv2-Tiny on ZOD.
    (a) Ground truth mask.
    (b) Simple averaging of RGB and LiDAR creates poorly defined boundaries.
    (c) Residual fusion recovers boundaries and suppresses structural artifacts.}
    \label{fig:ablation_fusion}
\end{figure*}

\begin{table}[t!]
\centering
\caption{Modality ablation (mIoU, \%). $\Delta_{\text{F-R}}$ is fusion minus RGB-only; $\Delta_{\text{R-L}}$ is RGB-only minus LiDAR-only.}
\label{tab:ablation_modality}
\resizebox{\columnwidth}{!}{%
\begin{tabular}{@{}ll|ccc|cc@{}}
\toprule
Dataset & Method & Fusion & RGB-Only & LiDAR-Only & $\Delta_{\text{F-R}}$ & $\Delta_{\text{R-L}}$ \\
\midrule
\multirow{13}{*}{ZOD}
 & CLFTv2-Large       & \textbf{53.54} & 51.63 & 38.27 & $+$1.91 & $+$13.36 \\
 & CLFTv2-Base        & 52.45 & 50.28 & 38.16 & $+$2.17 & $+$12.12 \\
 & CLFTv2-Tiny        & 43.44 & 40.79 & 30.54 & $+$2.65 & $+$10.25 \\
 & CLFT-Large         & 46.82 & 44.32 & 26.30 & $+$2.50 & $+$18.02 \\
 & CLFT-Hybrid        & 45.79 & 42.74 & 31.03 & $+$3.05 & $+$11.71 \\
 & CLFT-Base          & 44.64 & 41.95 & 23.72 & $+$2.69 & $+$18.23 \\
\cmidrule{2-7}
 & MaskFormer-Large   & 52.83 & 51.92 & 38.11 & $+$0.91 & $+$13.81 \\
 & MaskFormer-Base    & 52.34 & 51.39 & 37.66 & $+$0.95 & $+$13.73 \\
 & MaskFormer-Tiny    & 42.90 & 40.72 & 29.84 & $+$2.18 & $+$10.88 \\
 & Mask2Former-Large  & 52.52 & 51.28 & 38.45 & $+$1.24 & $+$12.83 \\
 & Mask2Former-Base   & 52.07 & 51.04 & 37.66 & $+$1.03 & $+$13.38 \\
 & Mask2Former-Tiny   & 43.28 & 40.59 & 31.36 & $+$2.69 & $+$9.23 \\
 & DeepLabV3+          & 36.70 & 29.67 & 23.73 & $+$6.03 & $+$5.94 \\
\midrule
\multirow{6}{*}{Waymo}
 & CLFT-Large         & \textbf{68.26} & 56.97 & 60.38 & $+$11.29 & $-$3.41 \\
 & CLFT-Base          & 66.32 & 55.02 & 58.26 & $+$11.30 & $-$3.24 \\
 & CLFT-Hybrid        & 65.50 & 56.53 & 59.09 & $+$8.97  & $-$2.56 \\
\cmidrule{2-7}
 & CLFTv2-Large       & 61.69 & 56.78 & 57.77 & $+$4.91 & $-$0.99 \\
 & CLFTv2-Base        & 61.02 & 56.13 & 56.88 & $+$4.89 & $-$0.75 \\
 & CLFTv2-Tiny        & 55.61 & 50.85 & 50.39 & $+$4.76 & $+$0.46 \\
\midrule
\multirow{13}{*}{ISEAuto}
 & CLFTv2-Large       & 73.30 & 72.96 & 63.92 & $+$0.34 & $+$9.04 \\
 & CLFTv2-Base        & 73.23 & 72.45 & 63.60 & $+$0.78 & $+$8.85 \\
 & CLFTv2-Tiny        & 69.94 & 68.73 & 59.11 & $+$1.21 & $+$9.62 \\
 & CLFT-Large         & 69.85 & 67.94 & 50.15 & $+$1.91 & $+$17.79 \\
 & CLFT-Hybrid        & 69.37 & 67.22 & 57.52 & $+$2.15 & $+$9.70 \\
 & CLFT-Base          & 65.67 & 64.13 & 35.51 & $+$1.54 & $+$28.62 \\
\cmidrule{2-7}
 & MaskFormer-Large   & \textbf{75.58} & 75.01 & 69.07 & $+$0.57 & $+$5.94 \\
 & MaskFormer-Base    & 75.21 & 74.80 & 69.09 & $+$0.41 & $+$5.71 \\
 & MaskFormer-Tiny    & 69.96 & 68.68 & 64.76 & $+$1.28 & $+$3.92 \\
 & Mask2Former-Large  & 75.37 & 74.87 & 68.39 & $+$0.50 & $+$6.48 \\
 & Mask2Former-Base   & 74.61 & 74.47 & 68.90 & $+$0.14 & $+$5.36 \\
 & Mask2Former-Tiny   & 70.13 & 68.69 & 65.90 & $+$1.44 & $+$2.79 \\
 & DeepLabV3+          & 56.97 & 51.92 & 47.46 & $+$5.05 & $+$4.46 \\
\bottomrule
\end{tabular}
}
\end{table}

\begin{table}[t!]
\centering
\caption{Progressive fusion ablation with CLFTv2-Tiny on ZOD.}
\label{tab:ablation_fusion}
\footnotesize
\setlength{\tabcolsep}{4pt}
\begin{tabular}{@{}p{0.54\columnwidth}cc@{}}
\toprule
Fusion Setting & mIoU (\%) & Time (ms) \\
\midrule
Add (RGB + LiDAR) & 23.20 $\pm$ 0.29 & 11.1 $\pm$ 0.4 \\
\midrule
Average ((RGB + LiDAR)/2) & 22.83 $\pm$ 0.36 & 10.8 $\pm$ 0.4 \\
\midrule
ResConv on both streams & 23.09 $\pm$ 0.36 & 13.3 $\pm$ 0.5 \\
\midrule
ResConv + residual from $M_{i-1}$ & \textbf{35.91 $\pm$ 0.47} & 13.2 $\pm$ 0.4 \\
\midrule
Residual + $\alpha$-gated variants (standard, LN, spatial, sigmoid) & 35.63 $\pm$ 0.34 & 13.1 $\pm$ 0.3 \\
\bottomrule
\end{tabular}
\end{table}

Table~\ref{tab:ablation_modality} shows that fusion consistently improves over RGB-only in all 32 settings ($\Delta_{\text{F-R}}>0$).
Waymo remains geometry-dominant, with LiDAR-only exceeding RGB-only in 5/6 transformer settings ($\Delta_{\text{R-L}}<0$), yet fusion still adds $+4.76\%$ to $+11.30\%$ over RGB-only.
Averaging the Waymo rows reveals a large architecture gap in fusion gain: CLFT achieves a mean $+10.52\%$ over RGB-only, while CLFTv2 reaches $+4.85\%$ (about $2.2\times$ smaller).
In contrast, CLFTv2 gains on ISEAuto are only $+0.34\%$ to $+1.21\%$, indicating limited marginal benefit from LiDAR when annotations are dense and the label space is simpler.
Figure~\ref{fig:ablation_modality} illustrates these complementary failure modes qualitatively.

Table~\ref{tab:ablation_fusion} shows that simple fusion rules (add, average, ResConv-only) cluster around 23\% mIoU, whereas adding residual propagation ($M_{i-1}$) lifts performance to 35.9\% over simple averaging (22.8\%).
This gain is accompanied by a moderate latency increase from 10.8 ms to 13.2 ms ($+22.2\%$), indicating a favourable accuracy-latency trade-off.
For the gated group, we evaluated four $\alpha$-gated variants: standard gated averaging, layer-normalized gated averaging, spatially weighted gated fusion, and sigmoid-constrained gated averaging.
In our current setup, gating mechanism did not improve mIoU, suggesting that the residual pathway already captures most of the useful cross-modal interaction under these conditions.
Figure~\ref{fig:ablation_fusion} provides a visual example of the boundary recovery enabled by residual fusion.

\subsection{Transfer Learning and Cross-Dataset Convergence}
\label{subsec:transfer}
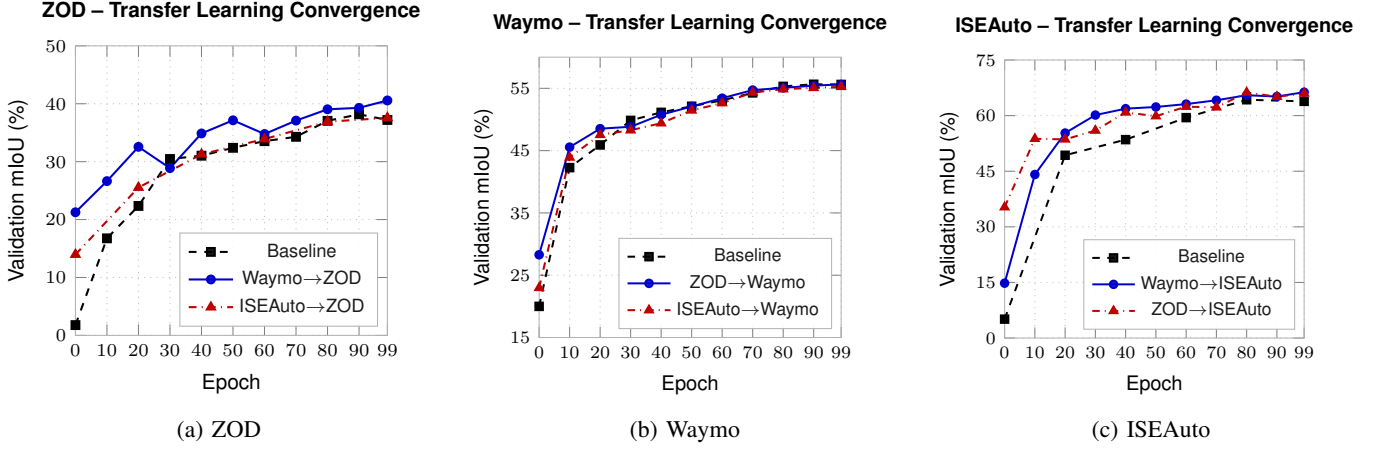
\begin{figure*}[t!]
    \centering
    \begin{subfigure}[b]{0.32\textwidth}
        \centering
        \resizebox{\linewidth}{!}{\begin{tikzpicture}
\begin{axis}[
    width=\linewidth,
    height=5.5cm,
    font=\sffamily\footnotesize,
    xlabel={Epoch},
    ylabel={Validation mIoU (\%)},
    xmin=0, xmax=99,
    ymin=0, ymax=50,
    xtick={0,10,20,30,40,50,60,70,80,90,99},
    ytick={0,10,20,30,40,50},
    grid=major,
    grid style={dotted, gray!50},
    legend pos=south east,
    legend style={font=\sffamily\scriptsize, draw=gray!60, fill=white, fill opacity=0.9},
    tick label style={font=\sffamily\scriptsize},
    label style={font=\sffamily\footnotesize},
    axis line style={gray!60},
    title style={font=\sffamily\footnotesize\bfseries},
    title={ZOD -- Transfer Learning Convergence},
    clip=false,
]

\addplot[
    color=black,
    dashed,
    thick,
    mark=square*,
    mark size=1.5pt,
    mark options={solid},
] coordinates {
    (0, 1.78) (10, 16.76) (20, 22.35) (30, 30.46) (40, 31.03)
    (50, 32.39) (60, 33.54) (70, 34.29) (80, 37.05) (90, 38.18) (99, 37.20)
};
\addlegendentry{Baseline}

\addplot[
    color=blue!80!black,
    solid,
    thick,
    mark=*,
    mark size=1.5pt,
] coordinates {
    (0, 21.26) (10, 26.64) (20, 32.56) (30, 28.88) (40, 34.87)
    (50, 37.14) (60, 34.77) (70, 37.09) (80, 39.04) (90, 39.29) (99, 40.57)
};
\addlegendentry{Waymo\ensuremath{\to}ZOD}

\addplot[
    color=red!75!black,
    dashdotted,
    thick,
    mark=triangle*,
    mark size=1.8pt,
    mark options={solid},
] coordinates {
    (0, 13.96) (20, 25.53) (40, 31.27) (60, 33.91) (80, 36.90) (99, 37.56)
};
\addlegendentry{ISEAuto\ensuremath{\to}ZOD}

\end{axis}
\end{tikzpicture}}
        \caption{ZOD}
        \label{fig:transfer_zod}
    \end{subfigure}
    \hfill
    \begin{subfigure}[b]{0.32\textwidth}
        \centering
        \resizebox{\linewidth}{!}{\begin{tikzpicture}
\begin{axis}[
    width=\linewidth,
    height=5.5cm,
    font=\sffamily\footnotesize,
    xlabel={Epoch},
    ylabel={Validation mIoU (\%)},
    xmin=0, xmax=99,
    ymin=15, ymax=60,
    xtick={0,10,20,30,40,50,60,70,80,90,99},
    ytick={15,25,35,45,55},
    grid=major,
    grid style={dotted, gray!50},
    legend pos=south east,
    legend style={font=\sffamily\scriptsize, draw=gray!60, fill=white, fill opacity=0.9},
    tick label style={font=\sffamily\scriptsize},
    label style={font=\sffamily\footnotesize},
    axis line style={gray!60},
    title style={font=\sffamily\footnotesize\bfseries},
    title={Waymo -- Transfer Learning Convergence},
    clip=false,
]

\addplot[
    color=black,
    dashed,
    thick,
    mark=square*,
    mark size=1.5pt,
    mark options={solid},
] coordinates {
    (0, 20.00) (10, 42.27) (20, 45.91) (30, 49.85) (40, 51.14)
    (50, 52.12) (60, 53.07) (70, 54.27) (80, 55.31) (90, 55.65) (99, 55.62)
};
\addlegendentry{Baseline}

\addplot[
    color=blue!80!black,
    solid,
    thick,
    mark=*,
    mark size=1.5pt,
] coordinates {
    (0, 28.28) (10, 45.56) (20, 48.52) (30, 48.83) (40, 50.76)
    (50, 52.08) (60, 53.41) (70, 54.72) (80, 55.07) (90, 55.44) (99, 55.62)
};
\addlegendentry{ZOD\ensuremath{\to}Waymo}

\addplot[
    color=red!75!black,
    dashdotted,
    thick,
    mark=triangle*,
    mark size=1.8pt,
    mark options={solid},
] coordinates {
    (0, 22.97) (10, 43.90) (20, 47.54) (30, 48.27) (40, 49.40)
    (50, 51.46) (60, 52.62) (70, 54.32) (80, 54.86) (90, 55.07) (99, 55.24)
};
\addlegendentry{ISEAuto\ensuremath{\to}Waymo}

\end{axis}
\end{tikzpicture}}
        \caption{Waymo}
        \label{fig:transfer_waymo}
    \end{subfigure}
    \hfill
    \begin{subfigure}[b]{0.32\textwidth}
        \centering
        \resizebox{\linewidth}{!}{\begin{tikzpicture}
\begin{axis}[
    width=\linewidth,
    height=5.5cm,
    font=\sffamily\footnotesize,
    xlabel={Epoch},
    ylabel={Validation mIoU (\%)},
    xmin=0, xmax=99,
    ymin=0, ymax=75,
    xtick={0,10,20,30,40,50,60,70,80,90,99},
    ytick={0,15,30,45,60,75},
    grid=major,
    grid style={dotted, gray!50},
    legend pos=south east,
    legend style={font=\sffamily\scriptsize, draw=gray!60, fill=white, fill opacity=0.9},
    tick label style={font=\sffamily\scriptsize},
    label style={font=\sffamily\footnotesize},
    axis line style={gray!60},
    title style={font=\sffamily\footnotesize\bfseries},
    title={ISEAuto -- Transfer Learning Convergence},
    clip=false,
]

\addplot[
    color=black,
    dashed,
    thick,
    mark=square*,
    mark size=1.5pt,
    mark options={solid},
] coordinates {
    (0, 5.10) (20, 49.31) (40, 53.52) (60, 59.46) (80, 64.28) (99, 63.85)
};
\addlegendentry{Baseline}

\addplot[
    color=blue!80!black,
    solid,
    thick,
    mark=*,
    mark size=1.5pt,
] coordinates {
    (0, 14.80) (10, 44.11) (20, 55.30) (30, 60.15) (40, 61.88)
    (50, 62.33) (60, 63.11) (70, 64.14) (80, 65.50) (90, 65.15) (99, 66.32)
};
\addlegendentry{Waymo\ensuremath{\to}ISEAuto}

\addplot[
    color=red!75!black,
    dashdotted,
    thick,
    mark=triangle*,
    mark size=1.8pt,
    mark options={solid},
] coordinates {
    (0, 35.29) (10, 53.75) (20, 53.59) (30, 55.97) (40, 60.87)
    (50, 59.85) (60, 62.41) (70, 62.23) (80, 66.24) (90, 65.10) (99, 65.98)
};
\addlegendentry{ZOD\ensuremath{\to}ISEAuto}

\end{axis}
\end{tikzpicture}}
        \caption{ISEAuto}
        \label{fig:transfer_iseauto}
    \end{subfigure}
    \caption{Transfer learning convergence for CLFTv2-Tiny on ZOD, Waymo, and ISEAuto (100 epochs total, sampled every 10).
    Initialisation from related dataset checkpoints systematically yields higher early-epoch mIoU compared to standard ImageNet pretraining.}
    \label{fig:transfer_convergence}
\end{figure*}

We evaluate cross-dataset transfer learning to determine if representations learned on one dataset generalize to others and accelerate training.
For each target dataset (ZOD, Waymo, ISEAuto) we train CLFTv2-Tiny both from ImageNet-pretrained weights (baseline) and from weights fine-tuned on each of the two other datasets (transfer).
Figure~\ref{fig:transfer_convergence} reports validation mIoU every 10 epochs across all conditions.
Transfer-initialised models consistently reach higher mIoU in early epochs across all three target datasets.
Class overlap mediates transfer quality: ISEAuto-to-ZOD transfer is weaker because ISEAuto lacks the sign class present in ZOD.
Initializing from a related dataset checkpoint accelerates convergence, which is especially beneficial for ZOD where learning from scratch is slow.

\subsection{Qualitative Analysis}
\label{subsec:qualitative}
\begin{figure*}[t!]
\centering
\begin{minipage}{0.16\textwidth}
\centering
\includegraphics[width=\textwidth, height=2.5cm]{images/zod/ground_truth/frame_000404.png} \\
ZOD GT
\end{minipage}
\hfill
\begin{minipage}{0.16\textwidth}
\centering
\includegraphics[width=\textwidth, height=2.5cm]{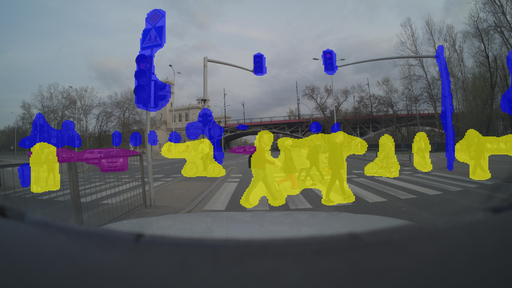} \\
ZOD CLFT
\end{minipage}
\hfill
\begin{minipage}{0.16\textwidth}
\centering
\includegraphics[width=\textwidth, height=2.5cm]{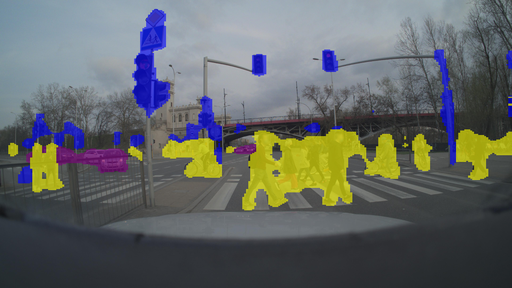} \\
ZOD CLFTv2
\end{minipage}
\hfill
\begin{minipage}{0.16\textwidth}
\centering
\includegraphics[width=\textwidth, height=2.5cm]{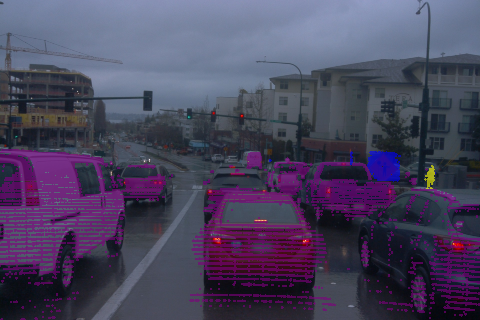} \\
Waymo GT
\end{minipage}
\hfill
\begin{minipage}{0.16\textwidth}
\centering
\includegraphics[width=\textwidth, height=2.5cm]{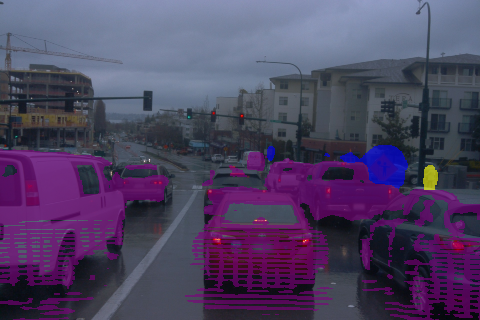} \\
Waymo CLFT
\end{minipage}
\hfill
\begin{minipage}{0.16\textwidth}
\centering
\includegraphics[width=\textwidth, height=2.5cm]{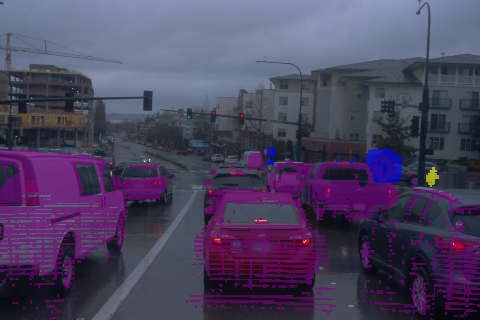} \\
Waymo CLFTv2
\end{minipage}

\vspace{0.5cm}

\begin{minipage}{0.16\textwidth}
\centering
\includegraphics[width=\textwidth, height=2.5cm]{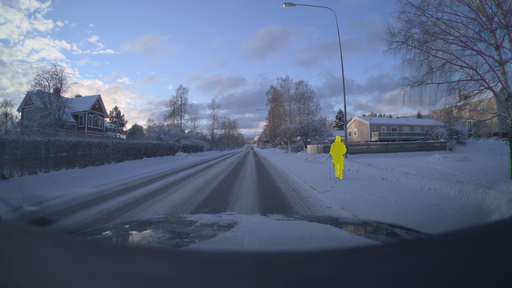} \\
ZOD GT
\end{minipage}
\hfill
\begin{minipage}{0.16\textwidth}
\centering
\includegraphics[width=\textwidth, height=2.5cm]{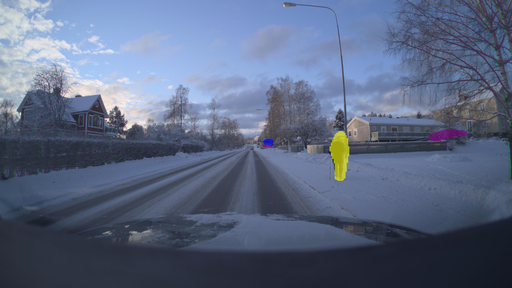} \\
ZOD CLFT
\end{minipage}
\hfill
\begin{minipage}{0.16\textwidth}
\centering
\includegraphics[width=\textwidth, height=2.5cm]{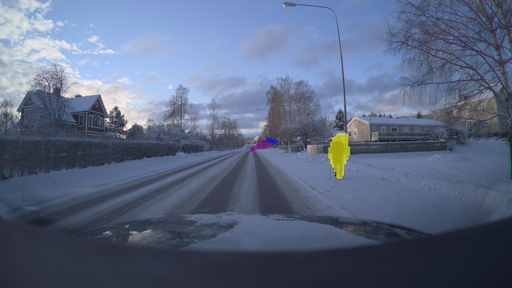} \\
ZOD CLFTv2
\end{minipage}
\hfill
\begin{minipage}{0.16\textwidth}
\centering
\includegraphics[width=\textwidth, height=2.5cm]{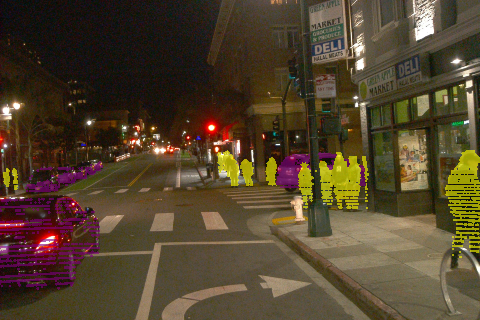} \\
Waymo GT
\end{minipage}
\hfill
\begin{minipage}{0.16\textwidth}
\centering
\includegraphics[width=\textwidth, height=2.5cm]{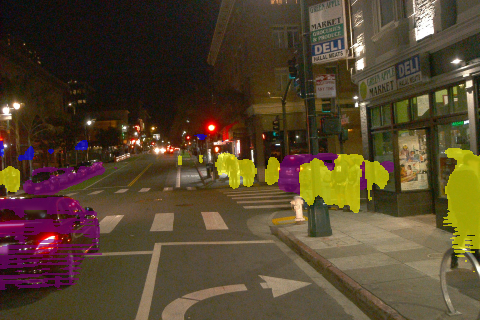} \\
Waymo CLFT
\end{minipage}
\hfill
\begin{minipage}{0.16\textwidth}
\centering
\includegraphics[width=\textwidth, height=2.5cm]{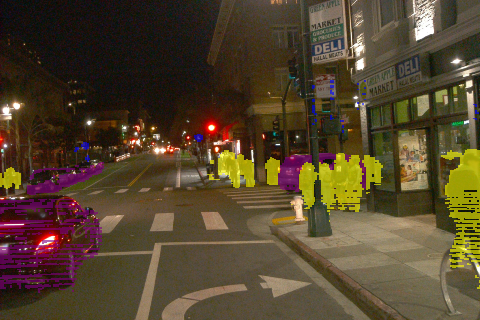} \\
Waymo CLFTv2
\end{minipage}

\caption{Qualitative comparison of CLFTv2-Large and CLFT-Hybrid models. Left: ZOD (Day Fair, Snow). Right: Waymo (Day Rain, Night Fair).}
\label{fig:qualitative_comparisons}
\end{figure*}

Figure~\ref{fig:qualitative_comparisons} compares representative predictions from CLFTv2-Large and CLFT-Hybrid across diverse ZOD and Waymo environments.

\subsection{Resilience and Efficiency Analysis}

\subsubsection{Safety-Critical Recall and Weather Resilience}
Table~\ref{tab:safety_metrics} shows a clear recall-precision trade-off.
CLFTv2-Large has the highest Waymo pedestrian recall (96.1\% Day Fair, 95.7\% Night Rain), while CLFT-Large has the highest Waymo precision (72.1\% Day Fair, 71.7\% Night Rain).
Table~\ref{tab:resilience_metrics} shows dataset-dependent difficulty: for CLFTv2-Base/Large, the Day Fair$\rightarrow$Night Rain mIoU drop is about 2.3--2.5\% on Waymo, 8.8--9.4\% on ZOD, and 15.9--17.1\% on ISEAuto.
On ZOD, CLFTv2-Large leads most weather splits, but Mask2Former-Base leads Night Rain (48.9\% vs 44.5\%).

\begin{table*}[t!]
\centering
\caption{Recall (\%) and precision (\%) for safety-critical classes under fusion. High recall minimizes missed objects (e.g., vulnerable road users); high precision limits false positives.}
\label{tab:safety_metrics}
\begin{tabular}{@{}l|cccc|cccc|cccc@{}}
\toprule
\multirow{3}{*}{Method} & \multicolumn{4}{c|}{ZOD} & \multicolumn{4}{c|}{Waymo} & \multicolumn{4}{c}{ISEAuto} \\
 & \multicolumn{2}{c}{Day Fair} & \multicolumn{2}{c|}{Night Rain} & \multicolumn{2}{c}{Day Fair} & \multicolumn{2}{c|}{Night Rain} & \multicolumn{2}{c}{Day Fair} & \multicolumn{2}{c}{Night Rain} \\
 & Rec. & Prec. & Rec. & Prec. & Rec. & Prec. & Rec. & Prec. & Rec. & Prec. & Rec. & Prec. \\
\midrule
CLFTv2-Large       & 83.5 & 51.4 & 82.5 & 52.1 & \textbf{96.1} & 64.5 & \textbf{95.7} & 65.4 & \textbf{98.6} & 75.9 & 90.9 & 60.1 \\
CLFTv2-Base        & \textbf{84.4} & 49.7 & \textbf{84.7} & 49.2 & 95.9 & 63.8 & 94.9 & 65.6 & 98.1 & 78.8 & 90.6 & 61.7 \\
CLFTv2-Tiny        & 78.1 & 39.5 & 83.2 & 42.4 & 94.9 & 57.3 & 86.8 & 55.8 & 97.6 & 72.3 & 90.7 & 57.0 \\
CLFT-Base          & 59.2 & 47.1 & 24.5 & 63.3 & 93.0 & 68.9 & 88.6 & 68.0 & 93.0 & 68.5 & 86.5 & 56.2 \\
CLFT-Hybrid        & 73.7 & 45.8 & 82.9 & 54.4 & 93.4 & 67.1 & 84.8 & 68.3 & 96.9 & 71.9 & 91.8 & 60.2 \\
CLFT-Large         & 59.1 & 49.3 & 39.2 & 64.8 & 92.5 & \textbf{72.1} & 87.2 & \textbf{71.7} & 97.5 & 72.8 & 85.9 & 63.1 \\
MaskFormer-Large   & 61.9 & \textbf{64.6} & 65.8 & \textbf{71.5} & 69.2 & 67.8 & 69.0 & 66.5 & 88.3 & 85.9 & 71.6 & 74.2 \\
MaskFormer-Base    & 62.9 & 63.9 & 67.3 & 68.2 & 66.5 & 69.9 & 64.2 & 65.6 & 88.3 & 86.4 & 71.5 & \textbf{75.5} \\
Mask2Former-Base   & 61.8 & 63.4 & 69.4 & 66.3 & 66.5 & 67.7 & 60.6 & 70.6 & \textbf{90.0} & 86.3 & 77.8 & 70.9 \\
Mask2Former-Large  & 61.4 & 64.1 & 68.7 & 63.8 & 66.0 & 67.9 & 65.9 & 70.7 & 89.1 & \textbf{87.0} & 77.5 & 75.0 \\
Mask2Former-Tiny   & 52.1 & 52.2 & 55.1 & 48.6 & 59.0 & 59.8 & 52.7 & 65.2 & 84.9 & 82.4 & 74.0 & 69.7 \\
MaskFormer-Tiny    & 47.5 & 57.4 & 50.8 & 56.3 & 59.8 & 60.8 & 56.9 & 56.7 & 83.5 & 82.0 & 74.8 & 68.4 \\
DeepLabV3+         & 75.2 & 28.6 & 74.5 & 29.3 & 92.5 & 49.0 & 77.3 & 45.1 & 97.9 & 61.4 & \textbf{94.1} & 42.1 \\
\bottomrule
\end{tabular}
\end{table*}
\begin{table*}[t!]
\centering
\caption{Weather Resilience Analysis on ZOD, Waymo, and ISEAuto fusion by Condition (mIoU \% and FW IoU \%). ISEAuto mIoU is the average of Human and Vehicle IoU (two classes; no traffic sign). Waymo and ISEAuto have no snow split.}
\label{tab:resilience_metrics}

\begin{tabular}{@{}llcccccccccccc@{}}
\toprule
Dataset & Method & \multicolumn{2}{c}{Day Fair} & \multicolumn{2}{c}{Day Rain} & \multicolumn{2}{c}{Night Fair} & \multicolumn{2}{c}{Night Rain} & \multicolumn{2}{c}{Snow} \\
\cmidrule(lr){3-4} \cmidrule(lr){5-6} \cmidrule(lr){7-8} \cmidrule(lr){9-10} \cmidrule(lr){11-12}
 &  & mIoU & FW IoU & mIoU & FW IoU & mIoU & FW IoU & mIoU & FW IoU & mIoU & FW IoU \\
\midrule
\multirow{13}{*}{\textbf{ZOD}}
 & CLFTv2-Large       & \textbf{53.3} & \textbf{61.8} & \textbf{55.8} & \textbf{66.7} & \textbf{52.6} & 56.6 & 44.5 & 44.6 & \textbf{53.7} & \textbf{68.4} \\
 & Mask2Former-Large  & 52.8 & 61.5 & 52.3 & 63.0 & 51.3 & \textbf{58.2} & 46.2 & 46.7 & 50.7 & 62.9 \\
 & MaskFormer-Large   & 52.5 & 60.6 & 53.4 & 63.5 & \textbf{52.6} & 58.1 & 45.8 & 44.5 & 53.2 & 65.8 \\
 & MaskFormer-Base    & 52.5 & 60.8 & 53.9 & 65.8 & 50.8 & 56.7 & 45.7 & 44.6 & 52.4 & 66.4 \\
 & Mask2Former-Base   & 52.4 & 61.2 & 53.5 & 64.4 & 50.4 & 56.3 & \textbf{48.9} & \textbf{49.8} & 52.6 & 65.1 \\
 & CLFTv2-Base        & 52.1 & 60.5 & 53.1 & 64.5 & 51.4 & 56.8 & 42.7 & 43.1 & 51.1 & 65.2 \\
 & CLFT-Large         & 46.2 & 55.6 & 46.5 & 59.7 & 41.9 & 53.3 & 35.4 & 37.0 & 48.2 & 58.7 \\
 & CLFT-Hybrid        & 46.2 & 55.0 & 48.5 & 60.4 & 48.4 & 56.3 & 39.7 & 37.8 & 45.6 & 55.9 \\
 & CLFT-Base          & 44.6 & 54.1 & 46.3 & 59.1 & 40.8 & 52.4 & 31.1 & 36.2 & 40.3 & 57.3 \\
 & Mask2Former-Tiny   & 43.0 & 52.2 & 42.8 & 55.4 & 41.6 & 49.8 & 33.1 & 33.6 & 44.2 & 58.1 \\
 & MaskFormer-Tiny    & 43.0 & 51.9 & 43.3 & 54.5 & 41.9 & 48.6 & 33.3 & 33.6 & 41.8 & 55.3 \\
 & CLFTv2-Tiny        & 42.7 & 52.1 & 44.3 & 55.6 & 43.8 & 52.5 & 34.1 & 33.8 & 42.3 & 58.0 \\
 & DeepLabV3+         & 33.8 & 43.5 & 34.8 & 47.7 & 31.7 & 40.4 & 26.3 & 27.8 & 32.1 & 49.3 \\
\midrule
\multirow{13}{*}{\textbf{Waymo}}
 & CLFT-Large         & \textbf{68.8} & \textbf{79.2} & \textbf{67.5} & \textbf{78.5} & \textbf{66.9} & \textbf{78.0} & \textbf{64.7} & \textbf{74.3} & -- & -- \\
 & CLFT-Base          & 66.7 & 78.2 & 65.5 & 77.5 & 65.2 & 77.3 & 62.9 & 72.9 & -- & -- \\
 & CLFT-Hybrid        & 66.0 & 76.7 & 65.3 & 76.0 & 64.9 & 76.3 & 62.3 & 72.2 & -- & -- \\
 & CLFTv2-Large       & 61.9 & 67.5 & 60.8 & 65.0 & 61.1 & 68.3 & 59.6 & 63.7 & -- & -- \\
 & CLFTv2-Base        & 61.2 & 67.2 & 60.1 & 64.8 & 60.5 & 68.0 & 58.7 & 63.1 & -- & -- \\
 & CLFTv2-Tiny        & 55.8 & 63.9 & 54.7 & 61.1 & 55.7 & 64.8 & 51.5 & 58.5 & -- & -- \\
 & Mask2Former-Large  & 49.5 & 54.3 & 48.9 & 50.8 & 49.0 & 56.0 & 47.1 & 48.8 & -- & -- \\
 & Mask2Former-Base   & 49.4 & 54.1 & 48.6 & 50.5 & 48.5 & 55.6 & 45.8 & 47.5 & -- & -- \\
 & MaskFormer-Large   & 51.0 & 56.8 & 50.5 & 53.9 & 50.5 & 57.9 & 47.9 & 50.5 & -- & -- \\
 & MaskFormer-Base    & 51.0 & 57.0 & 50.6 & 54.0 & 50.4 & 57.9 & 46.7 & 49.8 & -- & -- \\
 & DeepLabV3+         & 48.5 & 60.6 & 46.9 & 57.6 & 47.7 & 61.3 & 43.6 & 54.2 & -- & -- \\
 & Mask2Former-Tiny   & 42.1 & 50.8 & 41.6 & 47.8 & 42.0 & 52.3 & 39.4 & 44.6 & -- & -- \\
 & MaskFormer-Tiny    & 43.0 & 52.0 & 42.3 & 48.9 & 42.8 & 53.1 & 38.8 & 45.0 & -- & -- \\
\midrule
\multirow{13}{*}{\textbf{ISEAuto}}
 & Mask2Former-Large  & \textbf{80.7} & 81.7 & 73.6 & 83.5 & \textbf{73.3} & 79.0 & 65.4 & 68.7 & -- & -- \\
 & Mask2Former-Base   & \textbf{80.7} & 81.6 & 73.4 & 83.4 & 72.5 & 79.7 & 63.1 & 66.7 & -- & -- \\
 & CLFTv2-Base        & 80.5 & \textbf{82.0} & \textbf{75.7} & \textbf{84.0} & 72.6 & \textbf{80.6} & 63.4 & 68.1 & -- & -- \\
 & MaskFormer-Large   & 80.0 & 81.5 & 72.6 & 83.2 & 70.0 & 78.8 & 62.5 & 67.0 & -- & -- \\
 & MaskFormer-Base    & 79.9 & 81.1 & 72.2 & 83.0 & 73.1 & 79.4 & 63.4 & 68.2 & -- & -- \\
 & CLFTv2-Large       & 77.6 & 78.9 & 72.6 & 81.3 & 70.3 & 78.3 & 61.7 & 66.1 & -- & -- \\
 & Mask2Former-Tiny   & 75.5 & 77.4 & 65.0 & 79.2 & 66.5 & 74.8 & 60.7 & 64.8 & -- & -- \\
 & MaskFormer-Tiny    & 74.7 & 76.8 & 66.8 & 78.7 & 66.0 & 74.0 & \textbf{66.0} & \textbf{74.0} & -- & -- \\
 & CLFTv2-Tiny        & 74.2 & 75.8 & 68.6 & 79.0 & 66.0 & 74.8 & 60.4 & 66.2 & -- & -- \\
 & CLFT-Large         & 73.5 & 74.5 & 68.0 & 77.5 & 65.0 & 74.3 & 61.2 & 64.7 & -- & -- \\
 & CLFT-Hybrid        & 72.3 & 73.3 & 64.2 & 76.4 & 65.0 & 73.9 & 61.7 & 65.8 & -- & -- \\
 & CLFT-Base          & 68.9 & 70.7 & 65.5 & 74.8 & 59.0 & 69.9 & 56.0 & 59.7 & -- & -- \\
 & DeepLabV3+         & 64.5 & 66.6 & 56.5 & 68.7 & 56.5 & 65.6 & 49.2 & 56.3 & -- & -- \\
\bottomrule
\end{tabular}

\end{table*}

\subsubsection{Efficiency and Architectural Trade-offs}
\label{subsubsec:complexity_tradeoff}
Table~\ref{tab:efficiency} shows that CLFTv2 offers a strong efficiency-accuracy balance.
At Tiny scale, CLFTv2-Tiny runs at 22.0 ms and 187 MB, versus 48.4 ms and 314 MB for Mask2Former-Tiny (2.2$\times$ faster, about 40\% lower memory), while remaining competitive on ZOD/Waymo in the main benchmark tables.
At larger scale, CLFTv2-Large is also lighter than Mask2Former-Large (204.4G vs 265.3G FLOPs) and faster (81.8 ms vs 118.2 ms).
DeepLabV3+ remains the fastest model overall, but with a clear accuracy gap relative to the best transformer-based fusion models.

\begin{table}[t!]
\centering
\caption{A100 GPU inference efficiency. Latency is averaged over 100 runs.}
\label{tab:efficiency}
\resizebox{\columnwidth}{!}{%
\begin{tabular}{@{}lcccc@{}}
\toprule
Method & Params (M) & FLOPs (G) & Time (ms) & Mem (MB) \\
\midrule
DeepLabV3+      & 118.7 & 44.7  &  19.1 &  461 \\
CLFTv2-Tiny     &  43.1 & 30.9  &  22.0 &  187 \\
CLFT-Base       & 115.5 & 185.3 &  25.4 &  466 \\
MaskFormer-Tiny &  64.2 & 17.3  &  25.8 &  281 \\
CLFT-Hybrid & 127.6 & 185.5 & 37.2 &  512 \\
Mask2Former-Tiny &  71.0 & 42.3  &  48.4 &  314 \\
CLFTv2-Base     & 102.6 & 119.0 &  54.3 &  495 \\
MaskFormer-Base & 145.6 &  97.2 &  62.9 &  680 \\
CLFT-Large      & 341.2 & 440.5 &  63.3 & 1328 \\
CLFTv2-Large    & 211.4 & 204.4 &  81.8 &  907 \\
Mask2Former-Base & 152.4 & 153.0 &  90.7 &  719 \\
MaskFormer-Large & 316.8 & 209.5 &  91.9 & 1335 \\
Mask2Former-Large & 323.6 & 265.3 & 118.2 & 1374 \\
\bottomrule
\end{tabular}
}
\end{table}

\subsection{Limitations and Future Work}
Our evaluation reveals several limitations of local-window multi-modal fusion.
First, the observed accuracy gap between CLFTv2 and classical ViT-based architectures on the geometrically dense Waymo dataset indicates a limitation of local receptive fields when processing dense point clouds.
While shifted-window attention yields high efficiency, its localized receptive field limits cross-modal integration under conditions of extreme point-cloud density.
Future work should explore hybrid architectures that embed sparse, long-range token mixing within hierarchical backbones.

Second, our evaluation inherently incorporates the challenges of varied supervision quality.
Models evaluated on the ZOD dataset were supervised via SAM-generated pseudo-labels.
Evaluating on high-quality, human-annotated datasets in the future will help separate the model's architectural capabilities from its resilience to noisy pseudo-labels.

Finally, while the hardware efficiency metrics presented demonstrate substantial gains over query-based decoders, they are presently bound to a high-end compute profile (NVIDIA A100).
Validating these lightweight architectures across lower-power edge devices and micro-controllers represents a next step for real-time vehicular deployment.

\section{Code Availability}
\label{sec:code_availability}
To facilitate reproducibility and further research in multi-modal perception, the full source code for CLFTv2, including the training pipeline and model definitions, is publicly available at: \url{https://github.com/taltech-av/paper-tvt2026-clftv2}.
This repository also contains the data preprocessing and conversion scripts for all three training datasets.
Training results can be found on \url{https://app.visin.eu/projects/clftv2}.

\section{Conclusion}
\label{sec:concl}

This work introduces CLFTv2, a hierarchical Swin-based camera–LiDAR fusion framework designed to overcome the computational bottlenecks of multi-modal perception for autonomous driving.
CLFTv2 achieves a favorable accuracy-efficiency trade-off compared to leading universal segmentation architectures, reducing latency and compute overhead while maintaining performance on safety-critical classes.

Our evaluation across three distinct autonomous driving datasets demonstrates that architectural complexity is not universally beneficial: under the sparse and noisy supervision typical of large-scale open datasets, lightweight residual fusion performs on par with computationally demanding query-matching decoders.
Furthermore, we find that segmentation safety profiles are heavily architecture-dependent; the proposed CLFTv2 family prioritises high recall for vulnerable road users, while precision varies by dataset sensor characteristics.

These findings show the practical trade-offs between local-window efficiency and global cross-modal alignment, guiding the design of scalable, real-time perception architectures.

\bibliographystyle{IEEEtran}
\bibliography{references}

\end{document}